%% file: main.tex
\documentclass[10pt,journal,compsoc]{IEEEtran}
\usepackage{amsmath,amsfonts}

\usepackage{array}
\usepackage{subcaption}
\usepackage{textcomp}
\usepackage{graphicx}
\usepackage{capt-of}
\usepackage{amsmath,amssymb,amsfonts,dsfont,bm,bbm,mathrsfs,pifont}
\usepackage[ruled,linesnumbered]{algorithm2e}
\usepackage{listings}
\usepackage{booktabs,multirow,adjustbox}
\usepackage{float}
\usepackage[table]{xcolor}
\definecolor{citeblue}{HTML}{0071bc}
\usepackage[pagebackref,breaklinks,colorlinks,citecolor=citeblue]{hyperref}
\usepackage[capitalize]{cleveref} 
\crefname{section}{Sec.}{Secs.}
\Crefname{section}{Section}{Sections}
\crefname{table}{Tab.}{Tabs.}
\Crefname{table}{Table}{Tables}
\crefname{figure}{Fig.}{Figs.}
\Crefname{figure}{Figure}{Figures}
\crefname{equation}{Eq.}{Eqs.}
\Crefname{equation}{Equation}{Equations}
\crefname{algorithm}{Algorithm}{Algorithms}
\usepackage{stfloats}
\usepackage{url}
\usepackage{verbatim}
\usepackage{graphicx}
\usepackage{cite}

\definecolor{myblue}{HTML}{1f77b4}
\definecolor{myred}{HTML}{d62728}
\definecolor{boxred}{HTML}{AD2F2D}
\definecolor{mygreen}{HTML}{2ca02c}
\definecolor{myorange}{HTML}{ff7f0e}
\definecolor{mybrown}{HTML}{964d22}
\definecolor{nblgreen}{HTML}{009B5C}
\definecolor{myfirstred}{HTML}{B85450}  % RGB values for #F8CECC
\definecolor{Light}{HTML}{e7f5fe}
\usepackage{utfsym}
\usepackage{wrapfig}

\definecolor{textpurple}{RGB}{0,0,0}
\newcommand{\revise}[1]{\textcolor{black}{#1}}
\newcommand{\revised}[1]{\textcolor{textpurple}{#1}}

\usepackage{soul}
\newcommand{\etc}{\textit{etc.}}
\newcommand{\eg}{\textit{e.g.}}
\newcommand{\ie}{\textit{i.e.}}
\newcommand{\etal}{\textit{et al.}}
\newcommand{\vs}{\textit{vs.}}

\begin{document}

\title{Practical Continual Forgetting  for \\ Pre-trained Vision Models}

\author{
        Hongbo~Zhao, 
        Fei~Zhu, 
        Bolin~Ni,
        Feng~Zhu,
        Gaofeng~Meng,~\IEEEmembership{Senior~Member,~IEEE}
        and~Zhaoxiang~Zhang,~\IEEEmembership{Senior~Member,~IEEE}
\IEEEcompsocitemizethanks{
\IEEEcompsocthanksitem This work was supported in part by the National Key R\&D Program of China (No. 2022ZD0116500), the National Natural Science Foundation of China (No. U21B2042, No. 62320106010), the 2035 Innovation Program of CAS, and the InnoHK program. (Corresponding author: Fei Zhu; Zhaoxiang Zhang.)
\IEEEcompsocthanksitem 
H. Zhao, B. Ni, G. Meng and Z. Zhang are with the State Key Laboratory of Multimodal Artificial Intelligence Systems, Institute of Automation, Chinese Academy of Sciences, Beijing 100190, China.
E-mail: \{zhaohongbo2022, zhaoxiang.zhang\}@ia.ac.cn,   %
\IEEEcompsocthanksitem 
H. Zhao, B. Ni, G. Meng and Z. Zhang are also with the University of Chinese Academy of Sciences, Beijing 100190, China.
\IEEEcompsocthanksitem 
F. Zhu and  G. Meng are also with the Centre for Artificial Intelligence and Robotics, Hong Kong Institute of Science \& Innovation, Chinese Academy of Sciences, Hong Kong 999077, China.
E-mail: zhfei2018@gmail.com 
\IEEEcompsocthanksitem 
F. Zhu is with  SenseTime Research, Beijing 100084, China.
}
}

\markboth{Journal of \LaTeX\ Class Files,~Vol.~14, No.~8, August~2021}%
{Shell \MakeLowercase{\textit{et al.}}: A Sample Article Using IEEEtran.cls for IEEE Journals}

\IEEEpubid{0000--0000/00\$00.00~\copyright~2021 IEEE}

\input{sec/0_abs}

\maketitle

\IEEEdisplaynontitleabstractindextext

\IEEEpeerreviewmaketitle

\input{sec/1_intro}

\input{sec/2_related_work}
\input{sec/3_problem_setting}
\input{sec/4_method}

\input{sec/5_experiment}
\input{sec/6_discussion}

\input{sec/7_conclusion}

\ifCLASSOPTIONcaptionsoff
  \newpage
\fi

\bibliographystyle{IEEEtran}
\bibliography{IEEEabrv,main.bib}

\end{document}

%% file: sec/0_abs.tex
\IEEEtitleabstractindextext{\begin{abstract}
For privacy and security concerns, the need to erase unwanted information from pre-trained vision models is becoming evident nowadays.
In real-world scenarios, erasure requests originate \emph{at any time} from both users and model owners, and these requests usually form a sequence.
Therefore, under such a setting, selective information is expected to be \emph{continuously} removed from a pre-trained model while maintaining the rest.
We define this problem as continual forgetting and identify three key challenges.
\textbf{(i)} For unwanted knowledge, efficient and effective deleting is crucial.
\textbf{(ii)} For remaining knowledge, the impact brought by the forgetting procedure should be minimal. 
\textbf{(iii)} In real-world scenarios,  the training samples may be scarce or partially missing during the process of forgetting.
To address them, we first propose \textbf{G}roup \textbf{S}parse \textbf{LoRA} (GS-LoRA).
Specifically, towards \textbf{(i)},  we introduce  \revise{Low-Rank Adaptation (LoRA)} modules to fine-tune the \revise{Feed-Forward Network (FFN)} layers in Transformer blocks for each forgetting task independently, and
towards \textbf{(ii)}, 
a simple group sparse regularization is adopted, enabling automatic selection of specific LoRA groups and zeroing out the others.
To further extend GS-LoRA to more practical scenarios, we incorporate \textit{prototype} information as additional supervision and introduce a more practical approach, \textbf{GS-LoRA++}. 
For each forgotten class, we move the logits away from its original prototype. For the remaining classes, we pull the logits closer to their respective prototypes.
% which is effective, parameter-efficient, data-efficient, practical and easy to implement.
We conduct extensive experiments on face recognition, object detection and image classification and demonstrate that our method manages to forget specific classes with minimal impact on other classes.
Codes have been released on~\url{https://github.com/bjzhb666/GS-LoRA}.
\end{abstract}
}

%% file: sec/1_intro.tex
\section{Introduction}\label{sec:intro}
\IEEEPARstart{A}{s} pre-trained models become larger nowadays, more training data is required.
These data are usually collected through various ways, such as the Internet, books, publicly available datasets, and manual labeling.
Within the vast amount of data, there is often erroneous or privacy-sensitive information and pre-trained models may learn from it.
For instance, the ImageNet Roulette project~\cite{crawford2021excavating, ImageNet78:online} shows models tend to be biased toward racist, misogynistic, and cruel \etc~
Furthermore, with increased public awareness of privacy protection and updated privacy regulations~\cite{regulation2018general,goldman2020introduction}, individuals are now demanding the removal of any privacy-related information immediately.
Therefore, practical model erasing techniques are required upon receiving a deletion request.
\begin{figure}[t]
  \centering
   \includegraphics[width=\linewidth]{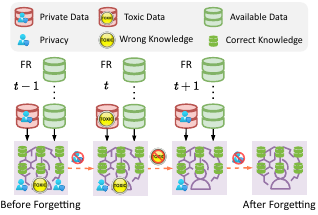}
   \caption{
   {Illustration of continual forgetting},
   which aims to remove specific knowledge in pre-trained models sequentially.
   ``FR'' stands for Forgetting Request.
   The \textcolor{myfirstred}{red} data (privacy data, toxic data, \etc) contains unwanted knowledge that needs to be removed, while the \textcolor{mygreen}{rest} should be maintained. 
   The model {inherits parameters} from the last forgetting task at the beginning of a new forgetting task.
   In practical scenarios, the forgotten and remaining data may be rare (few-shot), or some remaining data is missing.}
   \label{fig: motivation}
   \vspace{-10pt}
\end{figure}
In real-world scenarios,  these requests usually originate at any time from both users and model owners and naturally form a sequence. 
Under such a setting, selective information is expected to be \textit{continuously} removed from a pre-trained model while maintaining the rest. 
We identify this novel problem as \textbf{continual forgetting} and illustrate it in \cref{fig: motivation}, 
where privacy and wrong knowledge need to be removed from the pre-trained model sequentially.
This task holds significance in upholding privacy and reducing unwanted model bias, such as gender and racial discrimination, in real-world applications.

\revised{While the need for model erasure is broad, our primary focus is on functional forgetting, where the goal is to verifiably remove a model’s ability to perform a specific task (\eg, classify a particular category) for reasons such as regulatory compliance or model curation. For the granularity of forgetting, we concentrate on the class-wise setting. This approach serves as a standard and well-understood benchmark, providing a foundational step toward more complex instance-wise or identity-wise forgetting.} \revise{Besides, in practice, a critical challenge is that the data available to guide the forgetting process, both for the information to be removed and the information to be retained, is often scarce or missing. 
This transforms the problem into a few-shot or missing-data learning scenario, where naive fine-tuning can lead to overfitting, failing to properly erase the targeted knowledge while maintaining the rest.}

\begin{figure*}[t!]
    \centering
    \includegraphics[width=1\linewidth]{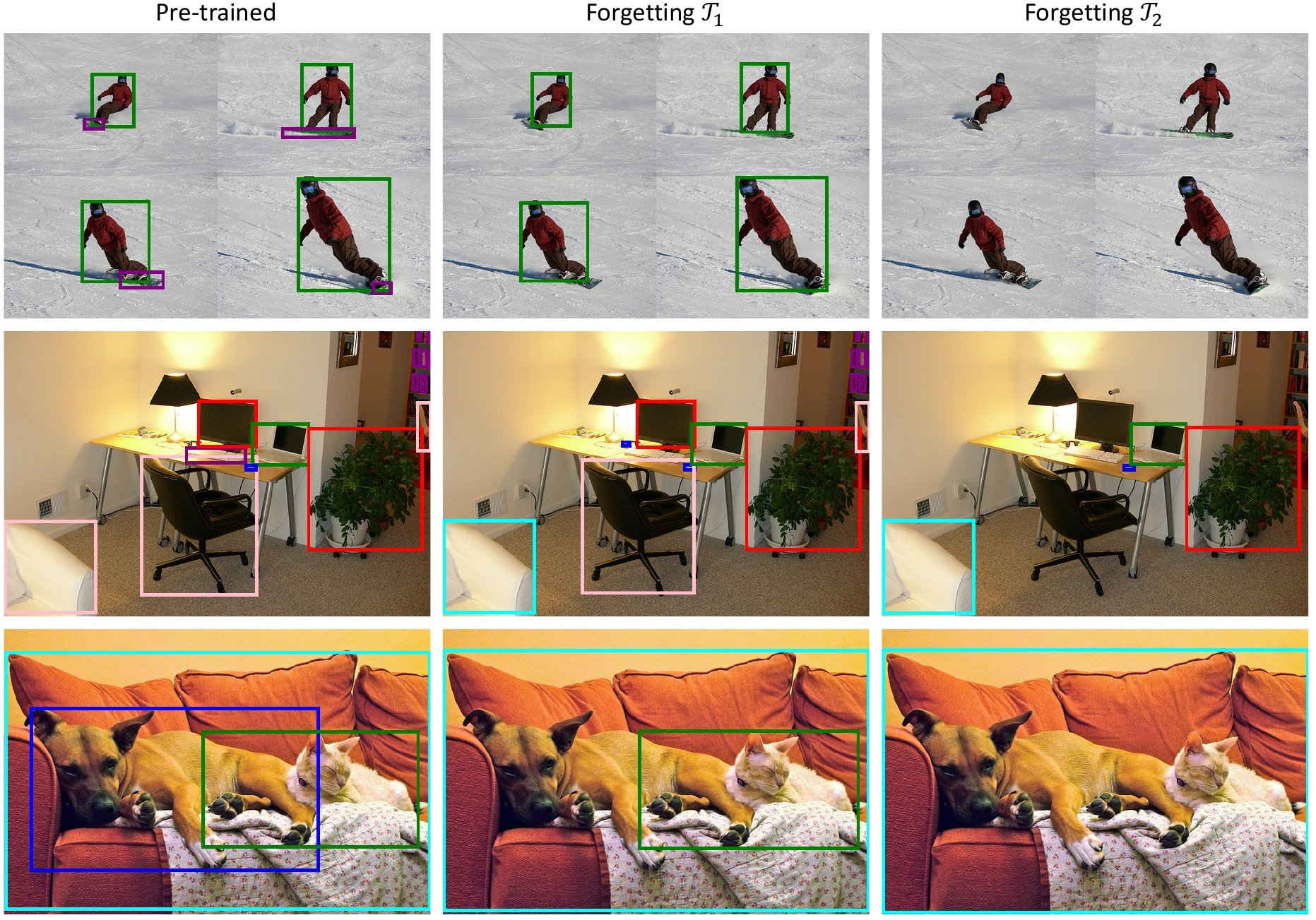}
    \caption{Visualization of {continual forgetting} on object detection tasks with COCO \cite{lin2014microsoft} dataset. 
    The left column (Pre-trained) shows the results from the pre-trained model.
    The middle column (Forgetting $\mathcal{T}_1$) shows the results when some classes (\eg, dog, keyboard, snowboard) are erased.
    The right column (Forgetting $\mathcal{T}_2$) shows the results when more objects (\eg, person, book, chair, cat) are erased.}
    \label{fig:cl-visual}
\end{figure*}
 
\begin{figure}[t]
    \centering
    \includegraphics[width=\linewidth]{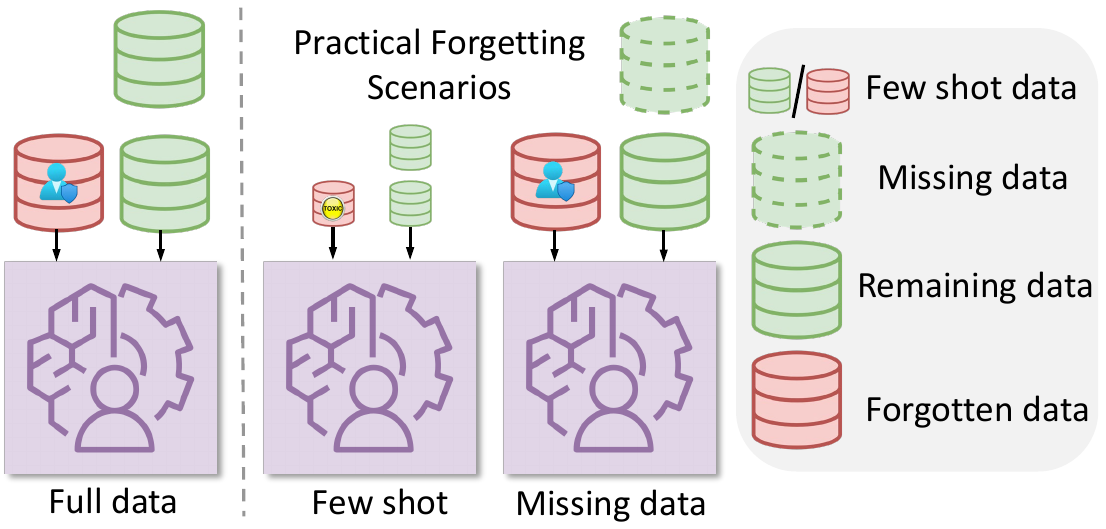}
    \caption{\revise{{Introductory figure of our scenarios.} The left part is one task in continual forgetting, \textit{i.e.}, one column in \Cref{fig: motivation}.
    }}
    \label{fig:scenarios}
    \vspace{-10pt}
\end{figure}

A related research topic is machine unlearning, which refers to the process of removing or erasing knowledge or patterns that a model has learned during training.
Prior attempts mostly focused on typical machine learning algorithms~\cite{mahadevan2021certifiable,bourtoule2021machine,chen2019novel,izzo2021approximate}, \eg, linear/logistic regression~\cite{mahadevan2021certifiable,izzo2021approximate}, and thus have a limited scope of application.
Recent studies that explored unlearning techniques for deep learning models are either computationally heavy and only effective on small-scale problems~\cite{guo2019certified,golatkar2020eternal,sekhari2021remember}, or require specific designs in the pre-training process~\cite{bourtoule2021machine, yan2022arcane}, which are impractical. 
These approaches lack the ability to proceed with numerous everyday requests and thus are not capable of practical continual forgetting.
In this work, we identify three key requirements in the design of practical continual forgetting algorithms.
\textbf{\textit{(i)}} Efficient and effective deleting for unwanted knowledge is crucial.
Especially for continual forgetting scenarios, lightweight and fast modifications are more important to achieve the deletion of information promptly.
\textbf{\textit{(ii)}} Should have minimal impact on remaining knowledge, \ie, catastrophic forgetting should be mitigated.
\textbf{\textit{(iii)}} Being robust in practical scenarios where the forgotten and remaining data may
be rare (few-shot) or some remaining data is missing.
\revise{We give an introductory figure of our scenarios in \Cref{fig:scenarios}.}

To this end, we first propose \textbf{G}roup \textbf{S}parse \textbf{LoRA} (GS-LoRA).
Specifically, to facilitate the forgetting of unwanted knowledge, we propose modifying the FFN modules within Transformer~\cite{vaswani2017attention} blocks, following Geva \etal~\cite{geva2020transformer}, who found that the Feed-Forward Network (FFN) modules in Transformer~\cite{vaswani2017attention} blocks store substantial amounts of knowledge.
% Specifically, on the one hand, to facilitate forgetting of unwanted knowledge, we propose modifying the FFN modules within Transformer~\cite{vaswani2017attention} blocks following Geva \etal~\cite{geva2020transformer} find that the Feed-Forward Network (FFN) modules in Transformer~\cite{vaswani2017attention} blocks store substantial amounts of knowledge. 
On the other hand, to realize \textit{efficient} forgetting, we introduce Low-Rank Adaptation (LoRA)~\cite{hu2021lora} to fine-tune the FFN modules inspired by parameter-efficient fine-tuning techniques~\cite{hu2021lora,houlsby2019parameter,li2021prefix}.
To mitigate catastrophic forgetting on remaining knowledge~\cite{kirkpatrick2017overcoming}, we propose a group sparse regularizer to enforce a sparse and accurate modification of FFN modules, as fine-tuning fewer parameters has been shown to  be effective~\cite{mallya2018piggyback, Mallya_Lazebnik_2018,l2p,zhang2022continual} in alleviating catastrophic forgetting.
This is akin to conducting minimally invasive surgery on a model instead of a major surgery.

To unleash the full potential of GS-LoRA, we further introduce {prototype} information and propose GS-LoRA++.
Specifically, GS-LoRA++ is designed to address more realistic and practical forgetting scenarios, where the samples for forgetting are rare due to the difficulty in acquiring the forgotten data.
This makes the forgetting process more challenging, and the model may not be able to fully forget the selected information.
Adding prototype guidance can effectively mitigate this issue and lead to satisfactory performance.
For each forgotten class, we shift the logits away from the original prototype, while for the remaining classes, we pull the logits closer to their respective prototypes.
GS-LoRA++ is practical, effective, parameter-efficient, data-efficient, easy to implement, and applicable to large models.

To verify the effectiveness of our proposed method, we conduct experiments on face recognition, image classification, and object detection tasks under single-step forgetting, continual forgetting, and practical forgetting scenarios.
Empirically, GS-LoRA++ effectively forgets specific classes while maintaining high performance on the remaining classes, indicating that our method is a general framework with minimal domain knowledge and few inductive biases across various vision tasks.
Qualitative results of object detection tasks on COCO \cite{lin2014microsoft} dataset are shown in \cref{fig:cl-visual}. It is observed that our method can attain selective removal without affecting the remaining classes.

\vspace{5pt} \noindent
\textbf{Extension of the conference version.}
This paper extends our conference version~\cite{zhao2024continual} in the following aspects.
First, we extend the proposed continual forgetting to more practical settings, including the few-shot setting and the missing class setting.
Second, we extend our method GS-LoRA \cite{zhao2024continual} with a novel prototype 
regularization, which can extract more information from the prototype of each class.
In this way, we can effectively alleviate the overfitting of the model, especially in the few-shot setting.
Third, we empirically show that our method can achieve effective and efficient forgetting on face recognition tasks, image classification, and object detection with minimal impact on the remaining classes in single-step forgetting, continual forgetting, and practical forgetting scenarios.
By introducing the practical continual forgetting problem and our simple and effective method, GS-LoRA++, we hope our findings can give the community an innovative and practical direction of continual learning and machine unlearning.

%% file: sec/2_related_work.tex
\section{Related Work}
In this section, we provide a brief introduction to the related literature, including continual learning, machine unlearning, parameter-efficient fine-tuning and few-shot learning.

\label{sec:related work}

\subsection{Continual Learning}

Continual learning aims to enable models to acquire new knowledge without forgetting previously learned information \cite{kirkpatrick2017overcoming}.
It is a learning paradigm that is particularly applicable in dynamic and changing scenarios.
Researchers have mainly designed four strategies to achieve this goal, including rehearsal-based methods~\cite{lavda2018continual,rebuffi2017icarl,isele2018selective,liu2023continual,rolnick2019experience,shin2017continual,zhu2021prototype,zhu2022learning,chen2023diffusepast}, regularization-based methods \cite{kirkpatrick2017overcoming,li2017learning,aljundi2018memory,schwarz2018progress,zhu2021class,zhu2024pass++}, \revise{structure-based methods~\cite{mallya2018piggyback,zhang2022continual,rusu2016progressive,Douillard_Rame_Couairon_Cord_2022,aljundi2017expert,liu2021adaptive,ni2024moboo,turner2024continual}} and prompt-based methods \cite{l2p,wang2022dualprompt,khattak2023self,wang2022s,smith2023coda}.
Rehearsal-based methods mitigate catastrophic forgetting by directly preserving samples from previous tasks \cite{rebuffi2017icarl,liu2023continual,feng2015learning,zhao2025mllm} or training an additional generative model to mimic past data \cite{shin2017continual,lavda2018continual}.
Regularization-based methods incorporate a regularization loss during the learning process of new tasks in order to limit the modification of important weights \cite{kirkpatrick2017overcoming,li2017learning,aljundi2018memory,schwarz2018progress,benjamin2018measuring} or employ the previous model as a teacher to supervise the current one \cite{rebuffi2017icarl, Wu_Chen_Wang_Ye_Liu_Guo_Fu_2019, Douillard_Cord_Ollion_Robert_Valle_2020}.
Structure-based methods either freeze specific architectures \cite{mallya2018piggyback, Mallya_Lazebnik_2018} to mitigate oblivion, or introduce newly designed structure growth modules \cite{rusu2016progressive,zhang2022continual,schwarz2018progress} to address new tasks.
Prompt-base methods \cite{l2p,wang2022dualprompt,khattak2023self,wang2022s,smith2023coda} use prompts to manage task-invariant and task-specific knowledge while maintaining model plasticity explicitly.
These strategies for continual learning are frequently combined together to improve performance \cite{rebuffi2017icarl,liu2023continual,zhu2023imitating,Ni_2024_CVPR,buzzega2020dark}.

Our proposed method falls into the category of structure-based methods.
However, our problem differs from continual learning as we aim to continuously delete, rather than add new knowledge to the model.

\subsection{Machine Unlearning}
\label{MUL}

\revise{Machine unlearning involves retraining or modifying machine learning models to diminish or eradicate the influence of previously acquired patterns or biases, aiming to enhance the models' fairness and safety \cite{nguyen2022survey, xu2023machine, bourtoule2021machine, shibata2021learning, brophy2021machine,ginart2019making,chundawat2023can,chundawat2023zero,kim2025we}.}
A lot of studies design unlearning algorithms on typical machine learning algorithms \cite{mahadevan2021certifiable,izzo2021approximate,baumhauer2022machine,chen2019novel,sun2023lazy,brophy2021machine}, like SVM \cite{chen2019novel}, linear model \cite{mahadevan2021certifiable}, random forests \cite{sun2023lazy}, \etc~
As a result, the applicability of these algorithms is constrained.
Initial work on forgetting in deep learning either slices the data and trains a series of submodels to isolate the effect of specific data points on the model \cite{bourtoule2021machine, yan2022arcane,shibata2021learning} (exact unlearning) or \revise{calculates influence functions to approximate the impact of a data item on the parameters of models \cite{golatkar2020eternal,guo2019certified,sekhari2021remember, jang2022knowledge, kurmanji2023towards,Ye2022LearningWR,liu2024model,cha2023learning,tarun2023fast,cha2024towards} (approximate unlearning).}
However, these methods deteriorate when applied to larger datasets and models, and the computational cost is exceedingly high.

Our problem focuses on the practical continual forgetting of a pre-trained model. 
One previous work \cite{shibata2021learning} studies the continual exact unlearning by adding a class-specific synthetic signal in the pre-training stage.
It should be noted that specific designs should not be performed in the pre-training process, which is not common in deep learning applications.
Practical continual forgetting requires algorithms to continually forget for any \textit{on-the-shelf} pre-trained models.

\subsection{Parameter-Efficient Fine-Tuning}

Training large models by self-supervised learning and then fine-tuning them on downstream tasks has become a new paradigm of deep learning \cite{gpt3,gpt2, He2022,he2020momentum,radford2021clip,li2023blip2,ni2022expanding}.
Parameter-efficient fine-tuning (PEFT) techniques \cite{hu2021lora,li2021prefix,houlsby2019parameter,jia2022visual,samadapter,zhou2022conditional, Tip-adapter} are proposed to optimize a \textit{limited} number of parameters, as fully fine-tuning increasing large models \cite{gpt2,gpt3,kirillov2023segment,MIR} becomes less practical for various downstream tasks.

Recent studies focus on three different types of PEFT methods, categorized based on the origin of trainable parameters. 
These methods include addition-based approaches \cite{li2021prefix, houlsby2019parameter, liu2021gpt}, which incorporate supplementary trainable neural modules or parameters; 
freezing-based techniques \cite{ravfogel2021bitfit,lee2019would},  which solely train a limited number of parameters from the original model; 
and parameter-factorization-based methods \cite{hu2021lora,valipour2022dylora, chavan2023one}, which utilize matrix low-rank factorization to update the model.
All these methods are designed to improve the performance of downstream tasks, while our method modifies pre-trained models with the help of PEFT.

\subsection{Few-shot Learning}
Few-shot learning \cite{ravi2017optimization,wang2020generalizing,sung2018learning,snell2017prototypical,sun2019meta,wang2023training,lu2023survey,fei2006one} aims to let a model learn to make accurate predictions by training on a very small number of labeled examples.
To achieve this goal, researchers have designed generative model-based approaches \cite{salakhutdinov2012one,fleuret2005pattern,wong2015one,lake2015human,fei2004learning,fei2006one} and discriminative model-based approaches \cite{lampert2013attribute,snell2017prototypical,hou2022closer,wang2018low,koch2015siamese,sung2018learning}.
Generative model-based approaches usually introduce some latent variables to describe the relationship between the input and output.
Most of these methods assume some specific distribution of the latent variables and use some typical machine learning methods like Bayesian estimation \cite{fei2006one,fei2004learning}, neural statistician \cite{edwards2016towards}, \etc~
Discriminative model-based methods utilize metric learning \cite{koch2015siamese,scott2018adapted,hertz2006learning,kaiser2017learning} and meta-learning strategies \cite{snell2017prototypical,hou2022closer,sung2018learning,finn2017model,ravi2017optimization,bertinetto2016learning}.
Metric learning methods create a continuous representation for a given data sample, such as a vector embedding.
They make inferences by learning a function that evaluates a distance metric, quantifying the similarity between this representation and the comparison of other samples or classes.
Meta-learning methods use prototypical network \cite{snell2017prototypical,hou2022closer}, matching network \cite{vinyals2016matching} or relation network \cite{sung2018learning}, \etc~to learn the ability to transfer across different tasks.
Another alternative solution to the problem is to generate or augment additional training samples \cite{kwitt2016one,dixit2017aga,yu2010attribute,gao2018low,wang2018low}.
These methods can be combined with generative model-based methods or discriminative model-based methods to improve the performance further.

%% file: sec/3_problem_setting.tex
\section{Problem Formulation}
% \color{textpurple}

\revise{In this section, we give the problem formulation of single-step forgetting, continual forgetting, and practical forgetting.
To improve clarity, we first define the notation used to formulate the problems in Tab. \ref{tab:notation}.}
% --- 表格代码 ---
\begin{table}[h]\color{black}
    \centering \small
\setlength{\tabcolsep}{2pt}
    \caption{\revise{Notation table of problem formulation.}}
    \begin{tabular}{ll}
    \toprule 
    Notation & Description \\ \midrule 
      $\mathcal{M}_0$   & The initial pre-trained model. \\
       $D_\text{train}$  & The original dataset used to train $\mathcal{M}_0$.\\
       $\mathscr{F}$ & The forgetting algorithm.\\
       $\mathcal{C}_{f}, \mathcal{C}_{r}$ & The set of classes to be forgotten and retained.\\
       $\mathcal{C}_\text{missing}$ & The set of remaining classes, but missing samples. \\
       $\mathcal{C}_\text{available}$ & The set of remaining classes with available samples. \\
       $\mathcal{T}$ & The total number of sequential forgetting tasks.\\
       $D_f^{(t)}$ & The forgotten dataset in task $t$.\\
       $D_r^{(t)}$ & The remaining dataset in task $t$.\\
       $D_f$ & The total forgotten dataset.\\
       $D_r$ & The total remaining dataset.\\
       $x_{f/r}^i$ & The $i$-th sample in the forgotten/remaining dataset.\\
       $y_{f/r}^i$ & The corresponding label of sample $x_{f/r}^i$. \\
       $\mathcal{M}_t$ & The updated model after forgetting task $t$.\\
       \revised{$\mathcal{L}$} & \revised{The origin loss function.} \\ \bottomrule
    \end{tabular}
    \label{tab:notation}
\end{table}
% \color{black}
\subsection{Single-step Forgetting}\label{sec:single-for}
We propose a new problem termed continual forgetting, which involves the selective removal of specific knowledge from a pre-trained model while preserving the performance of the rest.
In this subsection, we first consider the simplest situation where there is only one task that needs to be forgotten, and extend it to a continual form in \cref{sec:CL-forget}.
\revised{Let \revise{$\mathcal{M}_0$} be a model pre-trained on the dataset \revise{$D_\text{train}$}.}
% , we denote the mapping relationship of the model as 
% \revise{$f_{M_0}:\mathbb{X}_{D_\text{train}} \rightarrow \mathbb{Y}_{D_\text{train}}$, where $\mathbb{X}_{D_\text{train}}$ and $\mathbb{Y}_{D_\text{train}}$ represent the input set and output set, respectively.} 
Our objective is to selectively discard certain knowledge in the model while retaining the rest.
{Let $D_f$ and $D_r$ represent datasets containing knowledge to be forgotten and retained.
Given that $|D_r|$ is typically large in practical scenarios and the retraining process is time-consuming, we require $|D_r|+|D_f|\ll |D_\text{train}|$ for fast editing.}
Before forgetting, model \revise{$\mathcal{M}_0$} performs well on both {$D_f$ and $D_r$,} \revised{which is characterized by a low expected loss $\mathcal{L}$ on both datasets, \ie,} 
\revised{
\begin{equation}\small
    % \begin{cases}
    % \mathbb{E}_{(x,y)\sim D_f}L(f_{M_0}(x), y) \rightarrow \min \\
    % \mathbb{E}_{(x,y)\sim D_r}L(f_{M_0}(x), y) \rightarrow \min
    % \end{cases}.
    M_0=\arg\min \mathbb{E}_{(x,y)\sim D_f}L(f_{M_0}(x), y)\,\cap \,\arg\min \mathbb{E}_{(x,y)\sim D_r}L(f_{M_0}(x), y).
\end{equation}
}
The forgetting algorithm $\mathscr{F}$ modifies the model to obtain \revise{$\mathcal{M}_1 = \mathscr{F}(\mathcal{M}_0,D_f,D_r)$},
% \revised{The goal is for $\mathcal{M}_1$ to perform poorly on $D_f$ while maintaining performance on $D_r$. This is framed as the following optimization problem:}
\revised{by the following optimization:}

\revised{
\begin{equation}
\min_{M_1} \left[ \mathbb{E}_{(x,y)\sim D_r}L(f_{M_1}(x), y) - \lambda \, \mathbb{E}_{(x,y)\sim D_f}L(f_{M_1}(x), y)\right],
\end{equation}
}\revised{where $\lambda > 0$ is a hyperparameter balancing the two objectives. 
This formulation encourages the model to retain knowledge on $D_r$ and to forget knowledge on $D_f$.}

\subsection{Continual Forgetting} \label{sec:CL-forget}
Now, we extend the problem to a continual paradigm where the model is required to sequentially forget specific knowledge.
Let \revise{$D_r=\{D_r^{(t)}\}$ and $D_f=\{D_{f}^{(t)}\}$} for $t=1,2,\cdots, T$ represent two sequences of datasets, where $T$ is the number of forgotten tasks, $D_{f/r}^{(t)}$ is the forgotten or retained dataset of the $t$-th task.
For fast model editing, we still require that $|D_r|+|D_f|\ll |D_\text{train}|$.
The forgetting algorithm $\mathscr{F}$ handles erase requests sequentially, starting from \revise{$\mathcal{M}_0$, and generates a sequence of models $\mathcal{M}_{1},\mathcal{M}_{2},\cdots,\mathcal{M}_{t},\cdots,\mathcal{M}_{T}$, where $\mathcal{M}_{t}$ represents the modified model after the $t$-th forgetting task.
After processing task $t$, model $\mathcal{M}_{t}$ performs poorly on ${D_{f}^{(i)}}, (i=1,2,\cdots,t)$ but maintains the original performance in the remaining part.
\revised{This can be formulated as a single optimization problem:} \revised{
\begin{equation}\label{eq:clforget}\small
\min_{M_t} \left[ \mathbb{E}_{(x,y)\sim D_r^{(t)}}L(f_{M_t}(x), y) - \lambda \sum_{i=1}^{t} \mathbb{E}_{(x,y)\sim D_f^{(i)}}L(f_{M_t}(x), y) \right],
\end{equation} 
where $\lambda > 0$.
This ensures the model progressively erases knowledge from an expanding set of forgotten data while staying current on the retained data.}
}

\subsection{Practical Forgetting}
In the real-world scenario, the forgotten and retained data may contain few samples or be partially missing because of the storage.
Therefore, we consider two more practical and difficult settings from the data level, \ie, few-shot scenarios and missing class scenarios.
For simplicity, we only present the formulation in single-step forgetting as the task ID does not affect the structure of the formulation.

\vspace{5pt}\noindent
\textbf{Few-shot scenario.} In \cref{sec:single-for} and \cref{sec:CL-forget}, we require \revise{$|D_r|+|D_f|\ll |D_\text{train}|$}. 
Furthermore, we request a stricter setting where each class in $D_f$ and $D_r$ has a limited number $K$, \eg, 4 of samples, \ie,
\begin{equation}
    D_{{f}/{r}}=\{(x_{{f}/{r}}^i,y_{{f}/{r}}^i)\}_{i=1}^{K}.
\end{equation}

\vspace{5pt}\noindent
\textbf{Missing class scenario.}
In this setting, some classes in the remaining dataset have no training samples, \ie,
\begin{equation}\left\{
\begin{aligned}
     &D_{{f}}= \{(x_{{f}}^i,y_{{f}}^i) \}_{i=1}^{n_{f}} \\
     & \color{black}{D_r=D_{ra}+D_{rm}} \\
    &D_{{ra}}= \{(x_{{r}}^i,y_{{r}}^i) \}_{i=1}^{n_{r}}, \ y_{{r}}^i\in \mathcal{C}_\text{available}  \\
    &D_{{rm}}=\varnothing, \quad  \quad \quad \quad \quad y_{{r}}^i\in \mathcal{C}_\text{missing}
\end{aligned}\right.,
\end{equation}
where $D_{{ra}}$ is the available remaining dataset, $D_{{rm}}$ is the missing remaining dataset. $n_{f/r}$ are the total number of samples in forgotten/available remaining dataset.
The categories $ \mathcal{C}_\text{available}$ and $\mathcal{C}_\text{missing}$ constitute the entire categories $\mathcal{C}_r$ of the remaining dataset $D_r$.

%% file: sec/4_method.tex
\section{Method}

\begin{figure*}
    \centering
    \includegraphics[width=\linewidth]{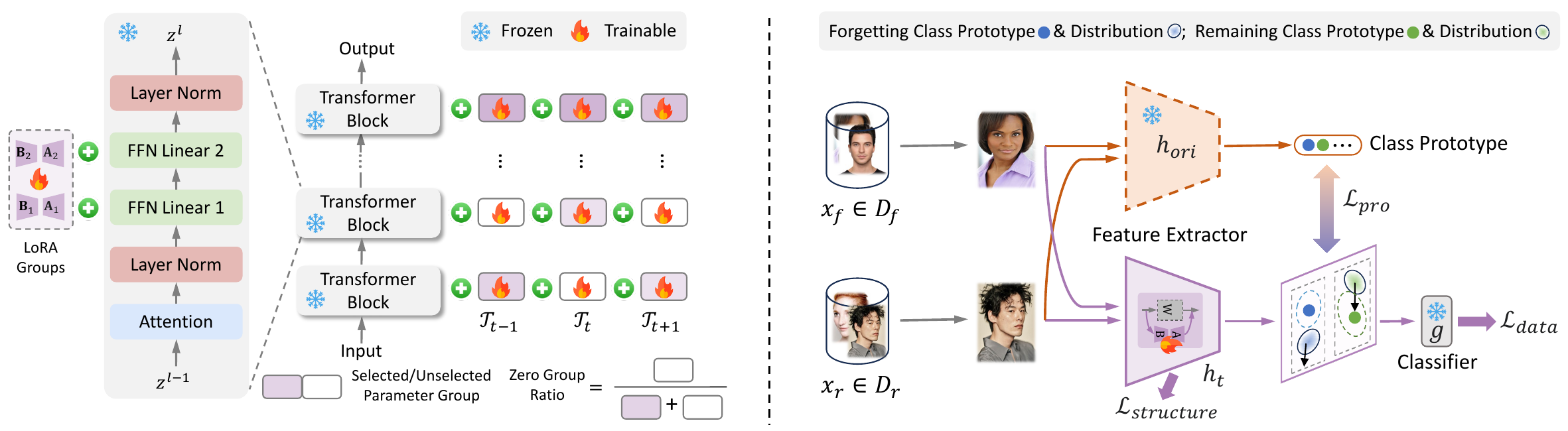}
    \caption{{Overall framework of GS-LoRA++.}
    We incorporate a set of LoRA modules in each continual forgetting task and propose a sparse structure selection strategy and prototype regularization to achieve accurate and few modifications.
    (Left) All LoRA modules are added in the Linear layers of FFN in the Transformer blocks, and we regard the LoRA modules in a Transformer block as one group.
    We use group sparse regularization ($\mathcal{L}_{\text{structure}}$) to automatically select LoRA groups.
    The purple groups are selected to modify and the white groups are neglected.
    The pre-trained model (including Transformer blocks and other parts) is frozen, and only LoRA groups are trainable.
    (Right) To achieve selective forgetting, we utilize selective forgetting and knowledge retention ($\mathcal{L}_{\text{data}}$).
    To further extend our method to more practical scenarios, we introduce prototype regularization $\mathcal{L}_{\text{pro}}$. 
    We use the original model to calculate the prototype of each class and \textit{pull away} logits from its original prototype for each forgotten class and \textit{pull in} logits from its own prototype for the remaining classes.
    }
    \label{fig:pipeline}
    \vspace{-0pt}
\end{figure*}

\textbf{Preliminary: LoRA.}    Hu \etal \cite{hu2021lora} argue that the weight matrix in the pre-trained model has a very low intrinsic rank and utilizes a low-rank decomposition to implement parameter updates.
For a weight matrix $\mathbf{W}\in \mathbb{R}^{d\times k}$, it is updated following $\mathbf{W} = \mathbf{W} + \Delta \mathbf{W} = \mathbf{W} + \mathbf{B}\mathbf{A}$, where $\mathbf{B}\in \mathbb{R}^{d \times r}$ and $\mathbf{A} \in \mathbb{R}^{r \times k}$ are low rank matrices and $r \ll min\{d,k\}$ is the rank of matrix $\mathbf{B}$ and $\mathbf{A}$.
Only matrices with low ranks, \ie, $\mathbf{B}$ and $\mathbf{A}$,  are trainable, while the matrix $\mathbf{W}$ remains frozen during training.
LoRA can be added to the linear projection matrices in Multi-Head Attention modules or the Feed-Forward Network (FFN) modules in Transformer \cite{vaswani2017attention} blocks.

\vspace{5pt} \noindent
\textbf{{Overview.}}
Considering three key challenges in \cref{sec:intro} and the optimization goal in \cref{eq:clforget}, we propose enhanced Group Sparse LoRA (GS-LoRA++) with selective forgetting, knowledge retention, and prototype regularization to achieve practical continual forgetting.
\cref{fig:pipeline} shows the overall pipeline of GS-LoRA++.
Specifically, to achieve efficient forgetting, we use LoRA to fine-tune the FFN modules in Transformer blocks. 
To mitigate catastrophic forgetting of the remaining knowledge, smaller network changes are preferred  \cite{Mallya_Lazebnik_2018,l2p,zhang2022continual,mallya2018piggyback}.
Therefore, we propose group sparse regularization to select and modify fewer blocks. 
\cref{GS-LoRA} gives a more detailed description.
To achieve the optimization goal in \cref{eq:clforget}, we use selective forgetting to maximize the original loss for the forgotten classes and knowledge retention to minimize the loss for the remaining classes in \cref{lossdata}.
Moreover, to achieve a practical forgetting, we introduce a prototype regularization to utilize more supervision from the prototypes in \cref{sec:prototypeloss}.
Finally, we provide the total learning objective and the pseudo-code in \cref{sec:final-object}.

\subsection{Low-rank Tuning with Group Sparsity}\label{GS-LoRA}

\textbf{LoRA Based Model Tuning.}
Following the findings of Geva \etal \cite{geva2020transformer}, FFN layers in the Transformer blocks store a substantial amount of knowledge, necessitating modification of the FFN modules to achieve knowledge erasure.
Although directly modifying these layers is theoretically feasible, it is inefficient due to the large number of parameters in the FFN layers.
To reduce the learnable parameters, we incorporate a set of LoRA modules into the FFN in each Transformer block and only make these LoRA modules trainable.

Suppose $\mathbf{x}$ is the input of the $l$-th FFN module, the mathematical form can be expressed as: 
\begin{equation}
    \text{FFN}^{(l)}(\mathbf{x}) = \max \left(\mathbf{0}, \mathbf{x}\mathbf{W}_1^{(l)} + \mathbf{b}_1^{(l)}\right)\mathbf{W}_2^{(l)} + \mathbf{b}_2^{(l)},
\end{equation} 
where $\mathbf{W}_1^{(l)}, \mathbf{W}_2^{(l)}, \mathbf{b}_1^{(l)}, \mathbf{b}_2^{(l)}$ are the weights and biases of two fully connected layers from the pre-trained model, respectively. 
We use LoRA to only fine-tune the weights of FFN modules:
\begin{equation}  
\begin{aligned}\label{eq:BA}
\mathbf{W}^{(l)}_t&=
\begin{bmatrix}
    {\mathbf{W}_1}^{(l)}_{t} \\
    {\mathbf{W}_2}^{(l)}_{t}
\end{bmatrix}=
\begin{bmatrix}
    \mathbf{W}_1^{(l)} \\
    \mathbf{W}_2^{(l)}
\end{bmatrix}+
\sum_{i=1}^{t} \mathbf{B}^{(l)}_i \mathbf{A}^{(l)}_i,\\
\mathbf{B}^{(l)}_i&=
\begin{bmatrix}
    {\mathbf{B}_{1}}^{(l)}_i & \mathbf{O} \\
    \mathbf{O} &   {\mathbf{B}_{2}}^{(l)}_i
\end{bmatrix},\quad
\mathbf{A}^{(l)}_i=
\begin{bmatrix}
     {\mathbf{A}_{1}}^{(l)}_i  \\
       {\mathbf{A}_{2}}^{(l)}_i
\end{bmatrix},
\end{aligned}
\end{equation}
where ${\mathbf{W}_1}^{(l)}_{t}$ and ${\mathbf{W}_2}^{(l)}_{t}$ denote the  weights of the $l$-th FFN modules after task $\mathcal{T}_t$, and  ${\mathbf{B}_1}^{(l)}_i,{\mathbf{A}_1}^{(l)}_i,{\mathbf{B}_2}^{(l)}_i,{\mathbf{A}_2}^{(l)}_i$ for $i=1,2,\cdots,t$ refer to the corresponding LoRA matrices in task $\mathcal{T}_i$.
$\mathbf{O}$ is the zero matrix.
Note that the output FFN layers are frozen to ensure forgetting occurs in the backbone and is difficult to recover. 
\cref{sec:6.1} provides detailed discussion.

\vspace{5pt}
\noindent\textbf{Group Sparsity Selection.}
To mitigate catastrophic forgetting and achieve precise modifications automatically, we introduce a group sparsity selection strategy that enables the selection of fewer Transformer blocks.
Although there are many ways to conduct a selection like routers \cite{yuksel2012twenty,shazeer2017outrageously}, meta learning \cite{vilalta2002perspective}, neural architecture search \cite{ren2021comprehensive,chen2019detnas}, 
we utilize \textit{group Lasso}, known for its simplicity and effectiveness in selecting parameters for specific groups \cite{yuan2006model, wen2016learning, feng2015learning,liu2015sparse} while setting others to zero.
Suppose LoRA matrices added to the $l$-th Transformer block in task $\mathcal{T}_t$ are ${\mathbf{B}_{1}}^{(l)}_t,{\mathbf{A}_{1}}^{(l)}_t,{\mathbf{B}_{2}}^{(l)}_t,{\mathbf{A}_{2}}^{(l)}_t$.
We regard the LoRA weights in one Transformer block as a group. Therefore, the group sparse loss in the $l$-th group can be written as:
\begin{equation}\label{eq:GS}
\mathcal{L}_{\text{gs}}^{(l)}=\Vert{\mathbf{B}}^{(l)}_t\Vert_\text{F}+\Vert{\mathbf{A}}^{(l)}_t\Vert_\text{F}.
\end{equation}
Here, $\Vert\cdot\Vert_\text{F}$ is the Frobenius norm of the LoRA matrices and $t$ denotes task $\mathcal{T}_t$.
Then, the group sparse loss on a set of weights can be represented as: 
\begin{equation}\label{eq:stru}
    \mathcal{L}_{\text{structure}} =\sum_{l=1}^G \mathcal{L}_{gs}^{(l)},
\end{equation} 
where $G$ is the number of groups, $\mathcal{L}_{gs}^{(l)}$ is the group sparse loss of the $l$-th group.%

\subsection{Selective Forgetting and Knowledge Retain} \label{lossdata}

\noindent
\textbf{Selective Forgetting.}
In each task $\mathcal{T}_t$ for $t=1,2,\cdots, T$, the model needs to forget the knowledge stored in data $D_{f_t}=(\mathbb{X}_{f_t},\mathbb{Y}_{f_t})$. 
To achieve forgetting,
the optimization goal is
$
    \textcolor{red}{\arg\max\limits_\mathbf{W}} \mathcal{L}\left(f_{M_{t-1}}(\mathbb{X}_{f_t}),\mathbb{Y}_{f_t}\right),
$
where $\mathbf{W}$ is the parameter;
$\mathcal{L}$ is the original loss function. 
An intuitive idea is to perform gradient ascent, \ie,
$\mathcal{L}_{\text{forget}}=-\mathcal{L}\left(f_{M_{t-1}}(\mathbb{X}_{f_t}),\mathbb{Y}_{f_t}\right)$.
Nevertheless, simply adding a minus sign to the original loss leads to an exploding \textit{unbounded} loss that is challenging to optimize. 
Therefore, we employ a ReLU function to introduce a lower bound following Du \etal \cite{du2019lifelong},
\begin{equation}\label{eq:forget}
\mathcal{L}_{\text{forget}}=\text{ReLU}\left(\text{BND}-\mathcal{L}\left(f_{M_{t-1}}\left(\mathbb{X}_{f_t}\right),\mathbb{Y}_{f_t}\right)\right),
\end{equation}
where BND is a hyperparameter that determines the bound.

\vspace{5pt} \noindent
\textbf{Knowledge Retention.}
Besides forgetting selected knowledge, it is crucial for the model to maintain performance on the rest. 
Catastrophic forgetting on remaining classes \cite{kirkpatrick2017overcoming} still exists.
To mitigate this issue, we employ a small rehearsal buffer $D_{r_t}=(\mathbb{X}_{r_t},\mathbb{Y}_{r_t})$ which satisfies $|D_{r_t}|+|D_{f_t}|\ll|D|$ to alleviate this undesirable forgetting and maintain efficient training. 
The knowledge retention loss can be written as:
\begin{equation}\label{eq:remain}
    \mathcal{L}_{\text{retain}}=\mathcal{L}\left(f_{M_{t-1}}\left(\mathbb{X}_{r_t}\right),\mathbb{Y}_{r_t}\right).
\end{equation}
Combining \cref{eq:forget,eq:remain}, we get the data loss
\begin{equation}\label{eq:dataloss}
    \mathcal{L}_{\text{data}} = \mathcal{L}_{\text{retain}} + \beta \mathcal{L}_{\text{forget}},
\end{equation}
where $\beta$ is a hyperparameter. 

\subsection{Prototype Regularization}\label{sec:prototypeloss}
% The information in one-hot labels is limited, and the network tends to overfit, especially in few-shot settings.
In practical scenarios, the training data for selective forgetting and knowledge retain may be rare, \ie, the forgetting task is under a few-shot setting.
% \todo{先讲few shot setting，再讲onehot}
\begin{table}[t!]
    \centering
    \caption{GS-LoRA\cite{zhao2024continual} does not effectively forget selective classes in few-shot settings due to overfitting.}
    \setlength{\tabcolsep}{9pt}
    \begin{tabular}{l|rrrr}
    \toprule
        \multirow{2}{*}{Dataset}  & \multicolumn{2}{c}{$Acc_f \downarrow $} & \multicolumn{2}{c}{$Acc_r \uparrow $} \\  \cmidrule(lr){2-3} \cmidrule(lr){4-5} 
           & Pre-train & Forget & Pre-train & Forget     \\ \midrule
        Training Set & 100.00\% &0.00\% &100.00\% &  100.00\%\\
        Test Set & 72.74\%  &\textcolor{red}{14.50\%} & 73.81\% &70.08\%\\
        % Remain Training Set &100.00\% &  100.00\%\\
        % Remain Test Set & 73.81\% &70.08\% \\
    \bottomrule
    \end{tabular}
    \label{tab:few_shot}
\end{table}
\Cref{tab:few_shot} shows the training accuracy and test accuracy when we forget 10 classes from a pre-trained Face Transformer under a 4-shot setting with GS-LoRA \cite{zhao2024continual}.
We can find that the model overfits the training samples easily \textit{without clearly forgetting the selective classes} (14.50\% in forget test set).
This is due to the limited information provided by the one-hot labels.
To solve this problem, we introduce prototype regularization to extract more information from the prototype.
We calculate the prototype $P_c$ \cite{snell2017prototypical,zhu2021class} of a certain class $c$ before forgetting as follows:
\begin{equation}\label{eq:pro-def}
    P_c=\frac{1}{N_c}\sum_{(x_i,y_i)\in S_c}h_{\text{ori}}(x_i),
\end{equation}
where $N_c$ is the number of samples from class $c$, $S_c$ is the set of class $c$, $h_{\text{ori}}(\cdot)$ denotes pre-softmax responses of the pre-trained model, \ie, logits.

\begin{algorithm}[t!]
\caption{GS-LoRA \& \textcolor{nblgreen}{GS-LoRA++}}\label{alg:gslora}
\KwIn{Pre-trained model weights $\theta_0$, number of forget tasks $T$, number of iteration $K$, original loss function $\mathcal{L}$, loss factors $\alpha,  \gamma,\beta$}
\KwData{Forgotten and retained data of $t$-th task $D_{{f}/r}^{(t)},~t=1,2,\cdots,T$}
\KwOut{Model with selective forgetting $T$ times}
\textcolor{nblgreen}{Calculate the prototype of each class using \cref{eq:pro-def}}\\
\For{i $\leftarrow$ 1 to T}
{
Freeze the model weights $\theta_{i-1}$\;
Add LoRA groups $\mathbf{B}_i^{(l)},~\mathbf{A}_i^{(l)}$ (\cref{eq:BA}) for each Transformer block\;
    \For{j $\leftarrow$ 1 to K}
    {
        Sample a mini-batch of forgotten and retained data\;
        Calculate the structure regularization using \cref{eq:GS,eq:stru}\;
        Calculate the selective forgetting using \cref{eq:forget}\;
       \textcolor{nblgreen}{ Compute the prototype regularization using \cref{eq:pro-total,eq:prof,eq:pror}}\;
       Compute gradient and update the parameters of $\mathbf{B}_i^{(l)},~\mathbf{A}_i^{(l)}$\;
       $j$++\;
    }
    Update the model weights $\theta_i=\theta_{i-1}+\mathbf{B}_i\mathbf{A}_i$\;
    $i$++\;
}
\end{algorithm}

To achieve selective forgetting, we destroy the prototypes of the forgotten classes and maintain the prototypes of the remaining classes.
Similar to the selective forgetting loss and the knowledge retention loss, we add two additional losses to constrain the prototype, \ie,
\begin{equation}\label{eq:prof}
    \mathcal{L}_{\text{pro}_f} = \text{ReLU}\left(\text{BND}-\text{KL}(P_f||h(x_f)) \right),
\end{equation}
where BND is a hyperparameter to determine the maximum Kullback-Leibler divergence \cite{kullback1997information} between the original prototype and the new logit of one sample $x_f$, $\text{KL}$ denotes the Kullback-Leibler divergence, $P_f$ is the original prototype of a certain forgotten class and $x_f$ is a sample from forgotten class $f$.

Correspondingly, for the remaining class, we maintain the logits similar to the prototypes using
\begin{equation}\label{eq:pror}
 \mathcal{L}_{\text{pro}_r} = \text{KL}(P_r|| h(x_r)), 
\end{equation}
where $P_r$ is the prototype of a certain remaining class $r$ and $h(x_r)$ denotes the logit of a sample.
Combining \cref{eq:prof} and \cref{eq:pror}, we get the prototype loss
\begin{equation}\label{eq:pro-total}
    \mathcal{L}_{\text{pro}} =  \mathcal{L}_{\text{pro}_r} + \gamma \mathcal{L}_{\text{pro}_f},
\end{equation}
where  $\gamma$ is a hyperparameter.

\begin{table*}[t!]
\centering
\caption{{Single-step forgetting results for face recognition.} $Acc_r\ (\%)$ and $Acc_f\ (\%)$ are the accuracies of remaining and forgotten classes.
$^*$~denotes the original methods with a rehearsal buffer.
Note that ``retrain'' represents retraining the model using replay data and \textit{the training epoch is the same as other methods to ensure a fair comparison.}
\revised{Retrain$^\dag$~means retraining the model using all replay data, not in the practical setting.}
Pre-train denotes the results before forgetting.
All settings are in the form of 100-Y, which means all experiments start from a pre-trained model (100 classes originally) and forget Y classes. 
For few-shot settings, we set 4 as the shot number.
Considering desired forgetting should consider $Acc_r$ and $Acc_f$ at the same time, we only  \textbf{bold} and \underline{underline} the best and second best $H$ results, respectively.
}
\vspace{-5pt}
\renewcommand{\arraystretch}{1.1}
\setlength{\tabcolsep}{6pt}
\begin{tabular}{lrrrrrrrrrrrrrr}
\toprule
\multirow{2}{*}{Methods} &\multirow{2}{*}{\begin{tabular}[c]{@{}c@{}}Tunable\\ Ratio $\downarrow$ \end{tabular}} & \multicolumn{3}{c}{100-5} & \multicolumn{3}{c}{100-10} & \multicolumn{3}{c}{100-50} & \multicolumn{3}{c}{100-90} \\ 
\cmidrule(lr){3-5} \cmidrule(lr){6-8} \cmidrule(lr){9-11} \cmidrule(lr){12-14} 
 && $H \uparrow$ & $Acc_r \uparrow$ & $Acc_f \downarrow$ & $H \uparrow$ & $Acc_r \uparrow$ & $Acc_f \downarrow$ & $H \uparrow$ & $Acc_r \uparrow$ & $Acc_f \downarrow$ & $H \uparrow$ & $Acc_r \uparrow$ & $Acc_f \downarrow$ \\ \midrule
Pre-train & - & - & 73.85 &  70.88& -  & 73.81 & 72.74 & - & 72.45 & 74.88 & - & 72.32 & 73.86 \\
Retrain$^\dag$ & - & - & \color{textpurple}71.69 &  \color{textpurple}0.00& -  & \color{textpurple}72.20 & \color{textpurple}0.00 & - & \color{textpurple}68.79 & \color{textpurple}0.00 & - & \color{textpurple}70.96 & \color{textpurple}0.00 \\ \hline
L2$^*$& 99.73\% & 67.13 & 66.96 & 3.58 & 67.74& 64.87&1.86& 64.39 & 56.61 & 0.24 & 52.45 & 40.68 & 0.04  \\
EWC$^*$ \cite{kirkpatrick2017overcoming} &99.73\%& 69.65 & 68.69 & {0.24} & 69.16 & 65.92 &{0.00}  & 62.48 & 53.61 & {0.00} & 44.41 & 31.75 & 0.00 \\
MAS$^*$ \cite{aljundi2018memory}&99.73\% & 69.73 & 69.29 & 0.72 & 69.38 & 66.41 & 0.12 & 62.67 & 53.88 & {0.00} & 46.79 & 34.24 &{0.00}  \\
LwF \cite{feng2015learning}&99.73\% & 67.95 &68.55&0.00 &70.08&67.62&0.00 &63.81&55.59&0.00 &61.25&52.32&0.00\\
DER \cite{buzzega2020dark}&99.73\% &69.70 &68.55&0.00 &70.08&67.62&0.00&63.81&55.59&0.00&57.03&46.44&0.00\\
DER++ \cite{buzzega2020dark}&99.73\% &69.70&68.56&0.00 &70.58&68.55&0.00&64.61&56.82&0.00&61.40&52.54&0.00\\
FDR \cite{benjamin2018measuring} &99.73\%&70.31&69.74&0.00 &70.03&67.51&0.00&65.40&58.04&0.00&53.85&42.37&0.00\\
SCRUB \cite{kurmanji2023towards}&99.73\%& 67.78 & 64.94 & {0.00} & 68.26& 64.31&{0.00}& 65.82 & 58.71 & 0.00 & 16.11 & 9.04 & {0.00}  \\
SCRUB-S \cite{kurmanji2023towards}&99.73\%& 70.29 & {69.95} & {0.24} & 71.19 & {69.81} &{0.12}  & 57.49 & 51.20 & {9.34} & 17.53 & 9.94 & {0.00} \\
LIRF$^*$ \cite{Ye2022LearningWR} &50.66\% & 25.56 & 67.67 & 55.13 & 26.35 & 65.83 & 56.26 & 47.13& 58.95 & {35.62} & 54.49 & 44.29 &{3.06}  \\
Retrain &100.00\%& 16.49&9.33&{0.00}&18.02&10.28&{0.00}&19.71&11.35&{0.00}&46.47&33.90&{0.00} \\
\rowcolor{Light}
GS-LoRA &\textbf{1.28\%}& \underline{71.02} & {71.16} & 0.00 & \underline{71.76} & {70.81} & {0.00}  & \underline{71.29} & {68.05} & 0.02 & \textbf{73.71} & {73.56} & 0.00 \\
\rowcolor{Light}
GS-LoRA++ &\textbf{1.28\%}& \textbf{71.12} & {71.36} & 0.00 & \textbf{72.04} & {71.35} & {0.00}  & \textbf{71.56} & {68.52} & 0.00 & \underline{72.85} & {71.86} & 0.00 \\ \midrule \multicolumn{14}{c}{\textit{Few-Shot Setting}} \\
\midrule
L2$^*$& 99.73\% & 39.56 & 67.83 & 42.96 & 50.99& 64.89&30.74& 50.18 & 39.53 & 6.20 & 7.51 & 3.95 & 0.00  \\
EWC$^*$ \cite{kirkpatrick2017overcoming} &99.73\%& 49.43 & 67.05 & {31.74} & 58.76 & 63.72 &{18.21}  & 46.69 & 34.23 & {1.48} & 4.39 & 2.26 & 0.00 \\
MAS$^*$ \cite{aljundi2018memory}&99.73\% & 49.68 & 67.27 & 31.50 & 58.23 & 62.94 & 18.56 & 45.84 & 33.31 & {1.40} & 4.39 & 2.26 &{0.00}  \\
LwF \cite{feng2015learning}&99.73\% & 58.50 &65.66&18.14 &58.27&64.68&19.72 &58.46&51.87&7.92 &34.61&22.60&0.00\\
DER \cite{buzzega2020dark}&99.73\% &64.25 &65.55&7.88 &\underline{64.09}&64.74&9.28&54.89&43.32&0.00&16.29&9.15&0.00\\
DER++ \cite{buzzega2020dark}&99.73\% &58.87&64.11&16.47 &59.44&63.73&17.05&\underline{61.42}&52.33&0.56&19.77&11.41&0.00\\
FDR \cite{benjamin2018measuring} &99.73\%&62.36&66.50&12.17 &62.08&66.31&14.39&61.07&51.78&0.45&25.75&15.59&0.00\\
SCRUB \cite{kurmanji2023towards}&99.73\%& 64.51 & 67.45 & {9.07} & 57.50& 65.44&{21.46}& 61.21 & 54.21 & 4.59 & \underline{37.19} & 25.08 & {1.96}  \\
SCRUB-S \cite{kurmanji2023towards}&99.73\%& 17.86 & {68.67} & {60.62} & 0.92 & {73.52} &{72.27}  & 12.06 & 66.55 & {68.25} & 0.00 & 0.00 & {1.99} \\
LIRF$^*$ \cite{Ye2022LearningWR} &50.66\% & 2.81 & 71.34 & 69.45 & 6.64 & 71.13 & 69.26 & 36.88& 56.52 & {47.51} & 27.44 & 16.95 &{1.85}  \\
Retrain &100.00\%& 0.28&0.14&{0.00}&0.49&0.25&{0.00}&0.51&0.25&{0.00}&0.00&0.00&{0.00} \\
\rowcolor{Light}
GS-LoRA &\textbf{1.28\%}& \underline{66.94} & {72.33} & 8.59 & {63.61} & {70.08} & {14.50}  & {4.72} & {74.88} & 72.45 & {26.82} & {16.38} & 0.00 \\
\rowcolor{Light}
GS-LoRA++ &\textbf{1.28\%}& \textbf{69.17} & {70.61} & 3.10 & \textbf{69.47} & {70.16} & {3.94}  & \textbf{70.61} & {67.01} & 0.26 & \textbf{60.86} & {51.75} & 0.01 \\
\bottomrule
\end{tabular}%
\label{tab:single-face}
\vspace{-0pt}
\end{table*}

\subsection{Final Learning Objective}\label{sec:final-object}
The final optimization goal can be expressed as follows:
\begin{equation}\label{loss}
    \mathcal{L}_{\text{total}} = \mathcal{L}_{\text{data}}+\mathcal{L}_{\text{pro}}+\alpha \mathcal{L}_{\text{strcuture}}.
\end{equation}
Here, $\mathcal{L}_{\text{data}}$ denotes the selective forgetting and knowledge retain and $\mathcal{L}_{\text{pro}}$ is the prototype regularization, which has been elaborated in \cref{lossdata} and \cref{sec:prototypeloss}, respectively. $\mathcal{L}_{\text{structure}}$ is the group sparse loss, and $\alpha$ serves as a hyperparameter to regulate the sparse intensity. 
We illustrate the final learning objective in \cref{fig:pipeline} (right) and give the pseudo-code of GS-LoRA and GS-LoRA++ in \cref{alg:gslora}.
The improvement of GS-LoRA++ is marked in \textcolor{nblgreen}{green}.

%% file: sec/5_experiment.tex
\section{Experiments}

\subsection{Experimental Setup} \label{sec5.1}

\textbf{Datasets and Pre-trained Models.}
We evaluate the effectiveness and efficiency of our methods using published Transformer-based models in face recognition, image classification and object detection tasks. 
For the face recognition task, we constructed a subdataset called CASIA-Face100 which collects 100 face IDs from the CASIA-WebFace \cite{yi2014learning} dataset.
We use a Face Transformer \cite{zhong2021face} pre-trained on the CASIA-Face100 dataset. 
For the image classification task, we use a standard VIT-B/16 \cite{dosovitskiy2020image} and load the pre-trained model from Pytorch \cite{paszke2019pytorch}.
The weights were trained from scratch by using a modified version of DeiT’s \cite{touvron2021training} training recipe.
We conduct forgetting on ImageNet100 \cite{russakovsky2015imagenet} dataset. 
For the object detection task, we use a deformable DETR \cite{zhu2020deformable} pre-trained on the COCO 2017~\cite{lin2014microsoft} dataset.%

\begin{table*}[t!]
\centering
\caption{{Continual forgetting results for face recognition.} $Acc_o \ (\%)$ is the accuracy of old tasks, \ie, the accuracy on all previously forgotten classes in task $\mathcal{T}_1,\mathcal{T}_2,\cdots,\mathcal{T}_{t-1}$.
There are 4 tasks in total and 20 classes are forgotten in each task.
For few-shot settings, we set 4 as the shot number.
Considering desired selective forgetting should consider $Acc_r$ and $Acc_f$ at the same time, we only  \textbf{bold} and \underline{underline} the best and second best $H$ and $Acc_o$ results respectively.}
\vspace{-5pt}
\renewcommand{\arraystretch}{1.1}
\setlength{\tabcolsep}{3.8pt}
\begin{tabular}{lrrrrrrrrrrrrrrr}
\toprule
\multirow{2}{*}{Methods} & \multicolumn{3}{c}{100-20} & \multicolumn{4}{c}{80-20} & \multicolumn{4}{c}{60-20} & \multicolumn{4}{c}{40-20} \\ \cmidrule(lr){2-4} \cmidrule(lr){5-8} \cmidrule(lr){9-12} \cmidrule(lr){13-16}
 & $H \uparrow$ & $Acc_r \uparrow$ & $Acc_f \downarrow$ & $H \uparrow$ & $Acc_r \uparrow$ & $Acc_f \downarrow$ & $Acc_o \downarrow$ & $H \uparrow$ & $Acc_r \uparrow$ & $Acc_f \downarrow$ & $Acc_o \downarrow$ & $H \uparrow$ & $Acc_r \uparrow$ & $Acc_f \downarrow$ & $Acc_o \downarrow$ \\ \midrule
Pre-train & - & 74.31 & 74.65 & - & 74.50 & 73.80 & - & - & 74.80 & 73.91 & - & - & 74.47 & 75.11 & - \\
\color{textpurple}Retrain$^\dag$ & - & \color{textpurple}74.53 & \color{textpurple}0.00 & - & \color{textpurple}73.29 & \color{textpurple}0.00 & \color{textpurple}0.00 & - & \color{textpurple}76.30 & \color{textpurple}0.00 & \color{textpurple}0.00 & - & \color{textpurple}74.69 & \color{textpurple}0.00 & \color{textpurple}0.00 \\ \hline
L2$^*$ & 66.91 & 62.13 & 2.16 & 66.66 & 61.74 & 1.37 & 9.36 & 66.42 & 61.37 & 1.54 & 11.83 & 66.95 & 61.02 & 0.97 &8.00 \\
EWC$^*$ \cite{kirkpatrick2017overcoming} &67.71  & 61.95 & {0.00} & 67.71 & 62.55 & {0.00} & \textbf{{0.00}} & 67.09 &61.43  &0.00  & {0.10} & 68.02 & 62.14 &0.00  &{0.23}  \\
MAS$^*$\cite{aljundi2018memory} & 67.52 & 61.63 & 0.00 & 68.12 & 63.25 & 0.00 & \textbf{{0.00}}& 68.15 & 63.23 &  {0.00}& \underline{0.03}   & 67.70 & 61.61 &{0.00} &\textbf{0.00} \\
LwF  \cite{feng2015learning}&69.43&65.04&0.21&69.94&66.47&0.00&0.21&70.34&67.11&0.00&\textbf{{0.00}}&70.89&67.12&0.00&\underline{0.04} \\
DER \cite{buzzega2020dark}&70.75&67.24&0.00&68.88&64.58&0.00&\textbf{{0.00}}&68.95&64.62&0.00&\textbf{{0.00}}&69.41&64.51&0.00&\textbf{0.00}\\
DER++ \cite{buzzega2020dark}&70.01&65.90&0.00&68.99&64.77&0.00&\textbf{{0.00}}&69.96&66.42&0.00&\textbf{{0.00}}&69.85&65.28&0.00&\textbf{0.00}\\
FDR \cite{benjamin2018measuring}&68.29&62.92&0.00&67.39&62.01&0.00&\textbf{{0.00}}&69.16&64.99&0.00&\textbf{{0.00}}&72.51&70.08&0.00&\textbf{0.00}\\
SCRUB  \cite{kurmanji2023towards}& 70.39 & 66.59 & {0.00} & 70.55 & 67.85 & {0.32} & \textbf{{0.00}} & 71.01 & 68.33 & {0.00} & {0.44} & 73.36 & 71.68 & 0.00 &{0.05} \\
SCRUB-S  \cite{kurmanji2023towards}&72.41  & 70.34 & {0.05} & {70.06}& {71.53} & {5.16} & {15.22} & {73.38}&73.09 &{0.24} & {15.81} & \textbf{75.33} & 76.24 &{0.68}  &{6.47}  \\
LIRF$^*$ \cite{Ye2022LearningWR}& 30.59 & 64.46 & 54.60 & 34.05 & 62.03 & 50.34 & {43.50}& 44.56 & 62.53 &  {39.29}& {36.58}   & 40.36 & 62.62 &45.34 &27.96 \\
Retrain &17.29&9.77&{0.00}&31.12&19.72&{0.00}&\textbf{{0.00}}&41.17&28.80&{1.72}&\textbf{{0.00}}&53.69&41.77&0.00&{0.09} \\
\rowcolor{Light}
GS-LoRA & \textbf{74.40} & {74.16} & 0.00 & \textbf{73.59} & {73.37} &  0.00& \underline{0.05} & \textbf{{74.36}} & {74.88} & 0.06 & \textbf{{0.00}} & {73.76} & {72.45} & 0.00 &{1.93}
\\ 
\rowcolor{Light}
GS-LoRA++ & \underline{73.97} & {74.16} & 0.00 & \underline{73.48} & {73.16} &  0.00& {0.10} & \underline{74.20} & {74.51} & 0.00 & \textbf{{0.00} }& \underline{75.23} & {75.36} & 0.00 &\textbf{0.00}\\
\midrule \multicolumn{16}{c}{\textit{Few Shot Setting}} \\
\midrule
L2$^*$ & 64.51 & 56.80 & 0.00 & 63.58 & 55.84 & 0.00 & 19.23 & 61.60 & 52.81 & 0.00 & 16.35 & 60.55 & 50.71 & 0.00 & 10.37 \\
EWC$^*$ \cite{kirkpatrick2017overcoming} &63.30 & 54.94 & 0.00 & 60.72 & 51.58 & 0.00 & 1.08 & 60.08 & 50.61 & 0.00 & 4.52 & 57.04 & 45.97 & 0.00 & 2.78  \\
MAS$^*$\cite{aljundi2018memory} & 63.49 & 55.24 & 0.00 & 61.11 & 52.14 & 0.00 & 1.13 & 58.79 & 48.81 & 0.00 & 1.40 & 56.11 & 44.79 & 0.00 & 0.40 \\
LwF  \cite{feng2015learning}&60.20 & 61.66 & 15.84 & 57.49 & 60.32 & 18.88 & 6.74 & 61.75 & 57.80 & 7.63 & 6.34 & 63.06 & 56.46 & 3.69 & 1.70 \\
DER \cite{buzzega2020dark}&\underline{66.97} & 60.72 & 0.00 & \textbf{66.09} & 59.83 & 0.00 & 0.93 & 64.30 & 56.90 & 0.00 & 2.94 & 59.22 & 48.87 & 0.00 & 0.40\\
DER++ \cite{buzzega2020dark}&65.87 & 58.94 & 0.00 & \underline{65.75} & 59.28 & 0.00 & \underline{0.31} & 63.42 & 55.54 & 0.00 & 2.11 & 62.93 & 54.15 & 0.00 & \underline{0.11}\\
FDR \cite{benjamin2018measuring}&65.53 & 61.37 & 4.37 & 62.05 & 60.67 & 10.31 & 1.34 & 63.96 & 58.58 & 3.49 & \underline{1.27} & 65.16 & 58.58 & 1.70 & 0.16\\
SCRUB  \cite{kurmanji2023towards}& 61.27 & 61.46 & 13.57 & 60.42 & 58.49 & 11.31 & 10.44 & 61.80 & 57.95 & 7.69 & 7.96 & 62.13 & 54.27 & 2.44 & 2.80 \\
SCRUB-S  \cite{kurmanji2023towards}&18.13 & 71.42 & 64.27 & 2.07 & 72.69 & 72.75 & 70.03 & 43.97 & 64.88 & 40.65 & 56.47 & 58.95 & 50.24 & 3.81 & 24.22  \\
LIRF$^*$ \cite{Ye2022LearningWR}& 11.59 & 69.50 & 68.33 & 23.09 & 64.91 & 59.76 & 59.43 & 30.56 & 61.25 & 53.55 & 53.07 & 34.73 & 57.17 & 50.17 & 47.20 \\
Retrain &0.54 & 0.27 & 0.00 & 1.58 & 0.80 & 0.00 & \textbf{0.00} & 3.07 & 1.57 & {0.00} & \textbf{0.00} & 11.99 & 6.52 & 0.00 &\textbf{ 0.00} \\
\rowcolor{Light}
GS-LoRA & 66.02 & 72.85 & 14.29 & 63.85 & 71.78 & 16.31 & 18.20 & \underline{67.25} & 70.53 & 9.64 & 15.47 & \underline{71.32} & 70.62 & 3.07 & 5.49 \\
\rowcolor{Light}
GS-LoRA++ & \textbf{71.24} & {68.25} & {0.15} & {65.73} & {69.07} & {11.10} & {0.93} & \textbf{68.03} & {67.26} & {5.09} & {6.03} & \textbf{72.10} & {69.91} & {0.68} & {4.03}\\
\bottomrule
\end{tabular}%

\label{tab:cl-face}
\end{table*}

\begin{table*}[t!]\color{black}
\centering
\caption{\color{black}{{Continual forgetting results on ImageNet100 \cite{russakovsky2015imagenet}.}
$Acc_m\ (\%)$ is the accuracy of the remaining classes missing training samples. 
}}
\vspace{-5pt}
\renewcommand{\arraystretch}{1.25}
\resizebox{1\textwidth}{!}{
\setlength{\tabcolsep}{2pt}
\begin{tabular}{lrrrrrrrrrrrrrrrrrrr}
\toprule
\multirow{2}{*}{Methods} & \multicolumn{4}{c}{100-20} & \multicolumn{5}{c}{80-20} & \multicolumn{5}{c}{60-20} & \multicolumn{5}{c}{40-20} \\ \cmidrule(lr){2-5} \cmidrule(lr){6-10} \cmidrule(lr){11-15} \cmidrule(lr){16-20}
 & $H \uparrow$ & $Acc_r \uparrow$ & $Acc_f \downarrow$ &$Acc_m \uparrow$  & $H \uparrow$ & $Acc_r \uparrow$ & $Acc_f \downarrow$ &$Acc_m \uparrow$ & $Acc_o \downarrow$ & $H \uparrow$ & $Acc_r \uparrow$ & $Acc_f \downarrow$ &$Acc_m \uparrow$ & $Acc_o \downarrow$ & $H \uparrow$ & $Acc_r \uparrow$ & $Acc_f \downarrow$ &$Acc_m \uparrow$ & $Acc_o \downarrow$ \\ \midrule
Pre-train  & -     & 89.93   & 87.40   & -     & -     & 90.03   & 89.60   & -     & -     & -     & 89.65   & 90.80   & -     & -     & -     & 92.80   & 86.50   & -     & -    \\
Retrain$^\dag$  & -     &\color{textpurple}82.31 &\color{textpurple}0.10   & \color{textpurple}0.10     & -     & \color{textpurple}83.02 &	\color{textpurple}0.00   & \color{textpurple}0.03     & \color{textpurple}0.00     & -     & \color{textpurple}82.35 &	\color{textpurple}0.00   & \color{textpurple}0.05    & \color{textpurple}0.10     & -     &\color{textpurple}88.50 	&\color{textpurple}0.00   & \color{textpurple}0.06     & \color{textpurple}0.01    \\
\hline
L2$^*$     & 84.03 & 81.18   & 0.30    & 45.28 & 85.79 & 82.37   & 0.10    & 4.42  & 2.10  & 86.82 & 83.85   & 0.80    & 2.24  & 4.80  & 85.95 & 85.40   & 0.00    & 0.28  & 5.97 \\
EWC$^*$      & 82.10 & 77.40   & 0.00    & 27.56 & 85.60 & 82.80   & 1.00    & 7.35  & 0.70  & 86.29 & 82.45   & 0.30    & 2.59  & 3.90  & 86.00 & 85.70   & 0.20    & 0.40  & 5.60 \\
MAS$^*$      & 81.17 & 75.85   & 0.10    & 27.55 & 80.57 & 73.27   & 0.10    & 1.52  & \underline{0.20}  & 82.06 & 74.85   & 0.00    & 0.22  & 1.20  & 80.14 & 74.80   & 0.20    & 0.10  & 1.50 \\
LwF$^*$      & 85.33 & 84.20   & 0.90    & 69.17 & \textbf{87.04} & 84.80   & 0.20    & 26.30 & 2.80  & \textbf{87.88} & 85.40   & 0.30    & 8.04  & 1.35  & 86.35 & 86.20   & 0.00    & 1.00  & 2.63 \\
DER       & 85.70 & 84.73   & 0.70    & 51.39 & 86.18 & 85.37   & 2.60    & 34.89 & 15.50 & 86.65 & 83.20   & 0.40    & 21.20 & 10.20 & 84.67 & 83.00   & 0.10    & 2.93  & 5.87 \\
DER++     & 85.72 & 84.68   & 0.60    & 56.45 & 86.10 & 84.57   & 1.90    & 33.05 & 12.60 & 86.65 & 83.20   & 0.40    & 22.82 & 12.10 & 84.51 & 82.60   & 0.00    & 2.15  & 6.07 \\
FDR       & 32.36 & 19.88   & 0.30    & 0.00  & 36.28 & 22.83   & 1.30    & 0.01  & 0.30  & 44.74 & 30.05   & 3.30    & 0.01  & 0.55  & 58.22 & 44.30   & 1.60    & 0.00  & 0.63 \\
SCRUB     & 76.06 & 67.33   & 0.00    & 0.22  & 75.97 & 65.93   & 0.00    & 0.00  & \textbf{0.00}  & 76.54 & 66.20   & 0.10    & 0.00  & \textbf{0.00}  & 75.38 & 66.80   & 0.00    & 0.00  & \textbf{0.00} \\
SCRUB-S   & 82.97 & 78.98   & 0.00    & 14.16 & 83.22 & 77.77   & 0.10    & 6.79  & \textbf{0.00} & 79.88 & 71.30   & 0.00    & 0.10  & \underline{0.05}  & 76.02 & 67.80   & 0.00    & 0.00  & \textbf{0.00} \\
Retrain   &  38.15   & 24.40  & 0.00    & 0.05  & 45.02 & 30.07   & 0.00    & 0.03  & \textbf{0.00}  & 49.90 & 34.40   & 0.00    & 0.07  & \textbf{0.00}  & 60.48 & 46.50   & 0.00    & 0.04  & \textbf{0.00} \\
\rowcolor{Light}
GS-LoRA   & \textbf{86.66} & {86.23}   & 0.30    & \underline{72.26} & \underline{86.86} & 84.47   & 0.20    & \textbf{59.56} & 0.40  & 87.60 & 84.70   & 0.10    & \underline{52.67} & 0.30  & \underline{88.17} & 89.90   & 0.00    & \textbf{34.18}& 0.07 \\
\rowcolor{Light}
GS-LoRA++ & \underline{86.49} & {86.38}   & 0.80    & \textbf{72.73} & 86.59 & 83.87   & 0.10    & \underline{57.64} & 0.40  & \underline{87.78} & 84.95   & 0.00    &\textbf{52.86} & 0.40  & \textbf{88.64} & 91.00   & 0.10    & \underline{30.63} & \underline{0.03}\\ \bottomrule
\end{tabular}
}

\label{tab:imagenet100}
\vspace{-5pt}
\end{table*}

\begin{table*}[t!]
\centering
\caption{{Continual forgetting results for face recognition with part of replay data missing.} We forget 20 classes per task on the face recognition task and 5 classes in the remaining dataset miss the training samples. $Acc_r\ (\%)$ is the accuracy of the remaining classes (including the class without training samples), $Acc_f\ (\%)$ is the accuracy of the forgotten classes, $Acc_o\ (\%)$ is the accuracy of old tasks, $Acc_m\ (\%)$ is the accuracy of the remaining classes missing training samples.
}
\vspace{-5pt}
\renewcommand{\arraystretch}{1.25}
\resizebox{1\textwidth}{!}{%
\setlength{\tabcolsep}{2pt}
\begin{tabular}{lrrrrrrrrrrrrrrrrrrr}
\toprule
\multirow{2}{*}{Methods} & \multicolumn{4}{c}{100-20} & \multicolumn{5}{c}{80-20} & \multicolumn{5}{c}{60-20} & \multicolumn{5}{c}{40-20} \\ \cmidrule(lr){2-5} \cmidrule(lr){6-10} \cmidrule(lr){11-15} \cmidrule(lr){16-20}
 & $H \uparrow$ & $Acc_r \uparrow$ & $Acc_f \downarrow$ &$Acc_m \uparrow$ & $H \uparrow$ & $Acc_r \uparrow$ & $Acc_f \downarrow$ & $Acc_o \downarrow$ &$Acc_m \uparrow$ & $H \uparrow$ & $Acc_r \uparrow$ & $Acc_f \downarrow$ & $Acc_o \downarrow$&$Acc_m \uparrow$ & $H \uparrow$ & $Acc_r \uparrow$ & $Acc_f \downarrow$ & $Acc_o \downarrow$ &$Acc_m \uparrow$\\ \midrule
 Pre-train&- & 73.50 & 74.45 & 70.92 & - & 73.22 & 74.28 & - & 70.92 & - & 72.94 & 73.79 & - & 70.92 & - & 73.34 & 72.56 & - & 70.92 \\ 
  \color{textpurple}Retrain$^\dag$ &- & \color{textpurple}61.64 & \color{textpurple}0.00 & \color{textpurple}0.00 & - & \color{textpurple}57.34 & \color{textpurple}0.00 & - & \color{textpurple}0.00 & - & \color{textpurple}57.48 & \color{textpurple}0.00 & - & \color{textpurple}0.00 & - & \color{textpurple}50.65 & \color{textpurple}0.00 & - & \color{textpurple}0.00 \\ 
  \hline
 L2$^*$&65.31 & 59.18 & 1.59 & 36.91 & 64.81 & 58.88 & 2.21 & 9.87 & 24.16 & 62.75 & 55.39 & 1.42 & 9.36 & 14.99 & 56.20 & 46.09 & 0.57 & 4.90 & 2.46 \\ 
 EWC$^*$ \cite{kirkpatrick2017overcoming}&66.47 & 60.04 & 0.00 & 25.95 & 64.57 & 57.10 & 0.00 & \textbf{0.00} & 4.92 & 62.99 & 54.99 & 0.06 & 0.10 & 0.67 & 57.47 & 47.57 & 0.00 & 0.09 & 0.00 \\ 
 MAS$^*$\cite{aljundi2018memory} &66.75 & 60.49 & 0.00 & 24.83 & 64.59 & 57.14 & 0.00 & \textbf{0.00} & 4.47 & 63.21 & 55.28 & 0.00 & \textbf{0.00} & 0.45 & 56.72 & 46.56 & 0.00 & \textbf{0.00} & 0.00 \\ 
 LwF  \cite{feng2015learning}&69.03 & 64.54 & 0.26 & 50.56 & 67.46 & 61.79 & 0.00 & \underline{0.05} & 28.19 & 66.34 & 60.30 & 0.06 & \textbf{0.00} & 12.75 & 59.70 & 50.71 & 0.00 & \textbf{0.00} & 2.24 \\ 
 DER \cite{buzzega2020dark}&67.71 & 62.10 & 0.00 & 15.66 & 66.47 & 60.14 & 0.00 & \textbf{0.00} & 0.67 & 63.56 & 55.83 & 0.00 & \textbf{0.00} & 0.00 & 59.86 & 50.95 & 0.00 & \textbf{0.00} & 0.00 \\ 
 DER++ \cite{buzzega2020dark}&67.35 & 61.49 & 0.00 & 9.62 & 65.60 & 58.74 & 0.00 & \textbf{0.00} & 0.67 & 64.53 & 57.34 & 0.00 & \textbf{0.00 }& 0.00 & 59.37 & 50.24 & 0.00 & \textbf{0.00} & 0.00 \\ 
 FDR \cite{benjamin2018measuring}&65.31 & 58.18 & 0.00 & 2.46 & 64.95 & 57.71 & 0.00 & \textbf{0.00} & 0.45 & 66.08 & 59.83 & 0.00 & \textbf{0.00} & 0.22 & 59.74 & 50.77 & 0.00 & \textbf{0.00} & 0.00 \\ 
 SCRUB  \cite{kurmanji2023towards}&64.82 & 57.47 & 0.10 & 9.40 & 65.93 & 59.26 & 0.00 & \textbf{0.00} & 0.45 & 65.64 & 59.11 & 0.00 & \textbf{0.00} & 0.00 & 61.68 & 53.67 & 0.06 & \textbf{0.00} & 0.00 \\
 SCRUB-S  \cite{kurmanji2023towards}& \underline{71.63} & 69.20 & 0.21 & 47.43 & 70.77 & 69.03 & 1.68 & 11.47 & 34.68 & 69.63 & 65.92 & 0.00 & 7.38 & 15.88 & 63.27 & 57.05 & 1.53 & 3.31 & 4.70 \\ 
 LIRF$^*$ \cite{Ye2022LearningWR}&30.97 & 63.65 & 53.98 & 58.84 & 34.62 & 60.78 & 50.08 & 42.11 & 43.40 & 43.94 & 57.48 & 38.22 & 31.80 & 32.89 & 36.04 & 53.55 & 45.40 & 24.28 & \textbf{28.64}\\
 Retrain &18.27 & 10.41 & 0.00 & 0.00 & 30.70 & 19.35 & 0.00 & \textbf{0.00} & 0.00 & 37.86 & 25.73 & 2.13 & \textbf{0.00} & 0.00 & 46.79 & 35.31 & 3.24 & 0.14 & 0.00 \\
 \rowcolor{Light}
 GS-LoRA &\textbf{72.13} & 70.05 & 0.10 & \textbf{63.09} & \textbf{72.37} & 70.55 & 0.00 & 0.15 & \textbf{58.61} & \underline{71.30} & 69.03 & 0.06 & \underline{0.03} & \textbf{49.44} & \underline{63.80} & 56.93 & 0.00 & \underline{0.02} & 12.53 \\
 \rowcolor{Light}
 GS-LoRA++ &71.15 & 68.13 & 0.00 & \underline{60.85} & \underline{71.33} & 68.61 & 0.00 & \textbf{0.00} & \underline{56.60} & \textbf{72.02} & 70.33 & 0.00 & \underline{0.03} & \underline{47.43} & \textbf{68.07} & 64.10 & 0.00 & \textbf{0.00} & \underline{23.94}\\
 \bottomrule
\end{tabular}
}

\label{tab:openset}
\vspace{-0pt}
\end{table*}

\vspace{5pt} \noindent
\textbf{Metrics.}
We need to evaluate the performance of the forgotten classes and the retained classes. 
We use the average accuracy (Acc) for classification tasks and the mean average precision (AP) for object detection tasks.
Ideally, the forgotten classes' performance should approach zero, and the remaining classes' performance should align with the original model's.
Similar to Shibata \etal \cite{shibata2021learning}, we define H-Mean to evaluate the overall performance after learning task $\mathcal{T}_t$, which is computed by:
\begin{equation}
    H\text{-}Mean^{(t)}=\frac{2Acc_r^{(t)} \cdot Drop^{(t)}}{Acc_r^{(t)}+Drop^{(t)}}.
\end{equation}
Here $Acc_r^{(t)}$ is calculated on the retained dataset after task  $\mathcal{T}_t$ and $Drop^{(t)}=Acc_f^{(origin)}-Acc_f^{(t)}$ is the performance drop on forgotten classes before and after training the task.\footnote{We take the classification problem as an example to define $H\text{-}Mean$. Replace $Acc$ with $AP$ for object detection tasks.}
After forgetting task $\mathcal{T}_t$, we evaluate the performance on all previously forgotten classes in task $\mathcal{T}_1,\mathcal{T}_2,\cdots,\mathcal{T}_{t-1}$ and denote it as $Acc_o$.

\subsection{Results and Comparisons}

\begin{figure*}[t!]
  \centering
    \begin{subfigure}{0.495\textwidth}
      \centering   
      \includegraphics[width=1\linewidth]{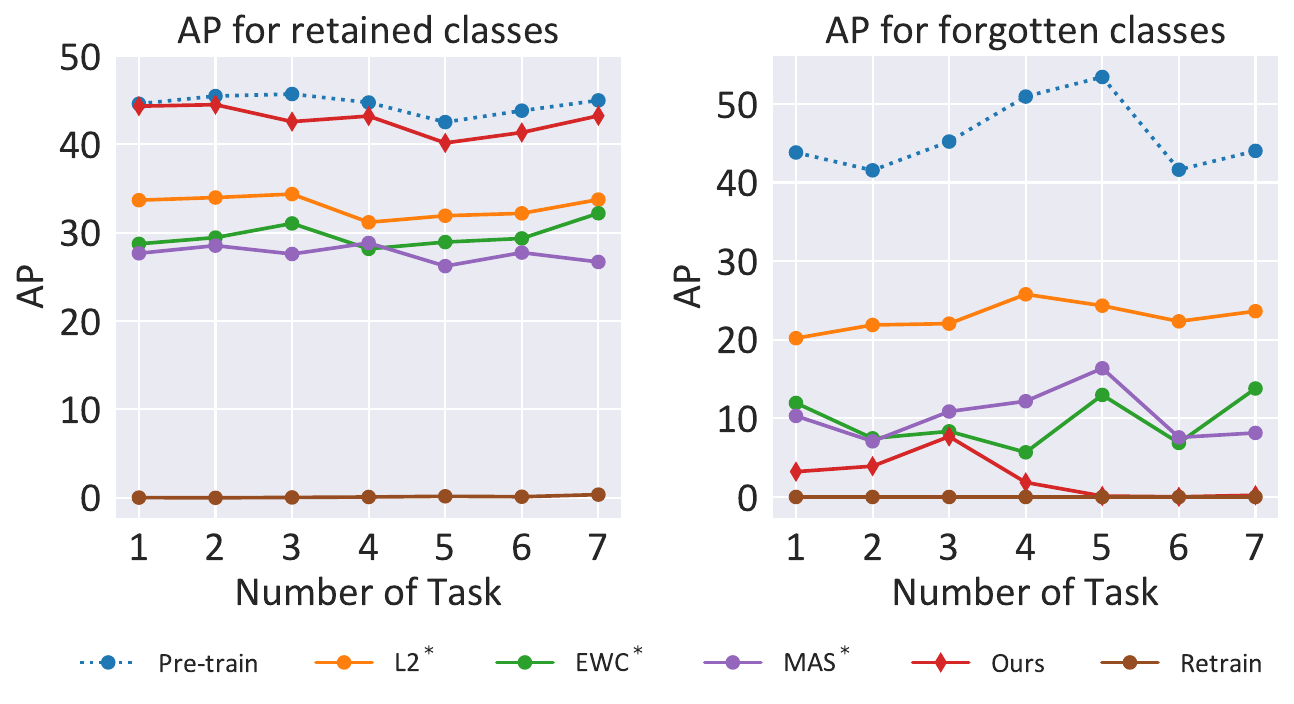}
        \caption{AP for retained  ($\uparrow$) and forgotten ($\downarrow$) classes.}
        \label{fig:sub1}
    \end{subfigure}   %
    \hfill  %
    \begin{subfigure}{0.495\textwidth}
      \centering   
      \includegraphics[width=\linewidth]{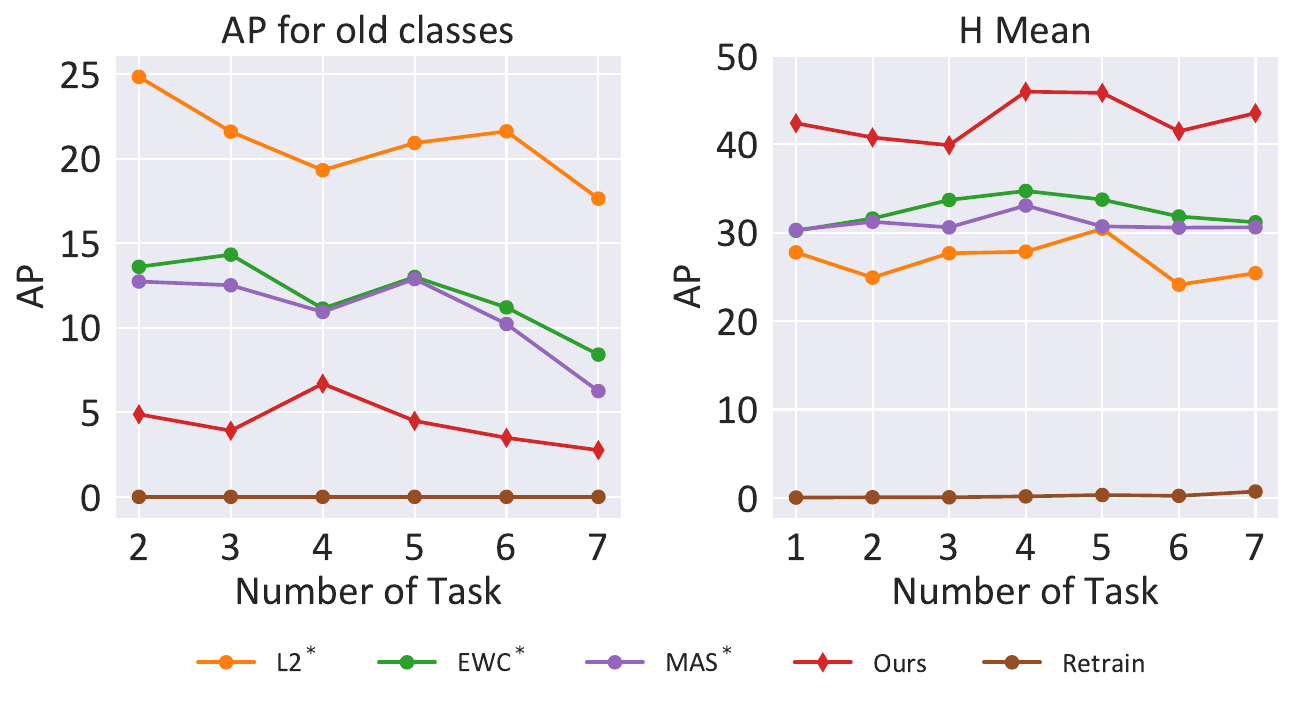}
        \caption{AP for old  ($\downarrow$) classes and H-Mean ($\uparrow$).}
        \label{fig:sub2}
    \end{subfigure}
    \vspace{-3pt}
\caption{
{Comparative results on object detection for continual forgetting.} Pre-train (\textcolor{myblue}{blue} lines) means the performance before forgetting; methods with * indicate the original methods with rehearsal buffer.
``Retrain'' (\textcolor{mybrown}{brown} lines) refers to the process of retraining the model using replay data and \textit{the training epoch is the same as other methods for a fair comparison.}
The \textcolor{myred}{red} line is our method. There are 7 tasks in total and 10 classes are forgotten in each task.
}
\label{fig:cl-obj}
\end{figure*}

We compared GS-LoRA and GS-LoRA++ with \textit{continual learning} methods including L2 regularization, EWC \cite{kirkpatrick2017overcoming}, MAS \cite{aljundi2018memory}, LwF~\cite{li2017learning}, DER \cite{buzzega2020dark}, FDR \cite{benjamin2018measuring}, \textit{machine unlearning} methods including LIRF \cite{Ye2022LearningWR}, SCRUB \cite{kurmanji2023towards} and retraining. 
Considering SCRUB~\cite{kurmanji2023towards} utilizes a min-max optimization method following GAN \cite{goodfellow2014generative}, we adopt a smoothing optimization method~\cite{zhang2020single} as a variant of SCRUB and name it SCRUB-S to further improve the performance.
Similar to our method, we freeze the final FFN layer to ensure backbone forgetting. 
For the retraining method, we use replay data to train a randomly initialized model and \textit{the training epoch is the same as other methods.}
We conduct extensive experiments on single-step, continual, and practical forgetting, including few-shot and missing class settings.

\noindent
\textbf{Single-step forgetting.}
\cref{tab:single-face} shows the performance comparisons with the aforementioned baselines for single-task forgetting on face recognition tasks, the degraded scenario in continual forgetting.
We consider to forget 5, 10, 50, 90 classes for a total of 100 classes.
The proposed GS-LoRA and GS-LoRA++ perform poorly in forgotten classes while retaining approximately the original performance in preserved classes. 
They are effective whether forgetting a small number of classes (\eg, 5 classes), or a large number of classes (\eg, 90\% of all the classes).
Besides, the tunable ratio of ours is much smaller than other methods, which shows the parameter-efficiency of our methods.

\noindent
\textbf{Continual forgetting.}
In continual forgetting scenarios, we conduct experiments on face recognition tasks, object detection tasks and image classification tasks.
For face recognition tasks, 20 classes are forgotten per task (4 tasks in total).
For image classification tasks, we use ImageNet100 \cite{russakovsky2015imagenet} dataset and let the model forget 20 classes per task (4 tasks in total).
For the object detection tasks, 10 classes are forgotten per task (7 tasks in total).

\cref{tab:imagenet100,tab:cl-face,fig:cl-obj} show the results for continual forgetting. 
Our method works the best among the listed methods across various tasks.
Meanwhile, our methods can achieve a relatively complete forgetting with a low old accuracy ($Acc_o$), while some baseline methods, \eg, DER and DER++~\cite{buzzega2020dark} in the image classification tasks, SCRUB-S~\cite{kurmanji2023towards} in face recognition tasks have a high undesired old accuracy.
These baselines may find some \textit{shortcuts} in the previous forgetting stage and the shortcut is restored in the current task, resulting in a high old accuracy.
Besides, we observe that in such a fast modification setting, severe underfitting occurs when using the retraining and LIRF \cite{Ye2022LearningWR}.
\revised{Although the soft upper bound retraining (Retrain$^\dag$ in the tables) achieves satisfactory performance, it requires 20x more training cost than our method.}

\begin{figure*}[t!]
  \centering
    \begin{subfigure}{0.495\textwidth}
      \centering   
      \includegraphics[width=1\linewidth]{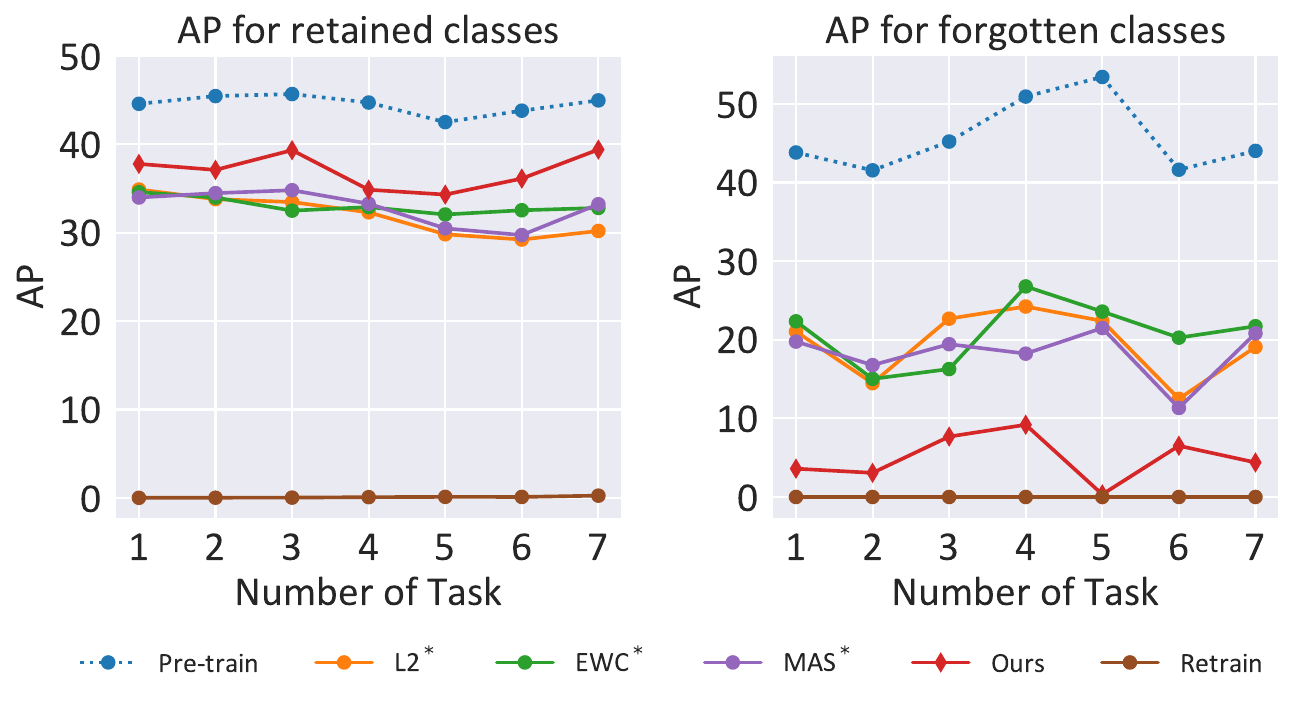}
        \caption{AP for retained  ($\uparrow$) and forgotten ($\downarrow$) classes.}
        \label{fig:fewsub1}
    \end{subfigure}   %
    \hfill  %
    \begin{subfigure}{0.495\textwidth}
      \centering   
      \includegraphics[width=\linewidth]{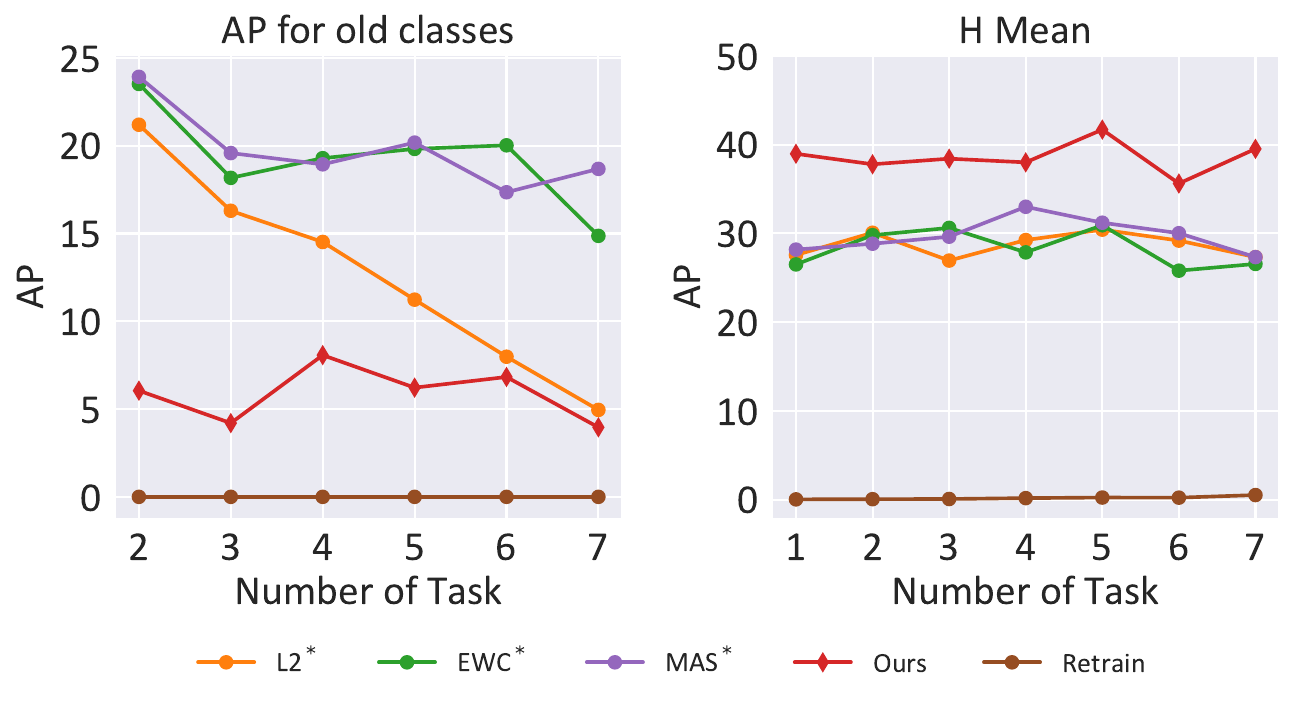}
        \caption{AP for old  ($\downarrow$) classes and H-Mean ($\uparrow$).}
        \label{fig:fewsub2}
    \end{subfigure}
    \vspace{-3pt}
\caption{
{Comparative results on object detection for continual forgetting under the few-shot setting.}
}
\label{fig:cl-obj-few}
\end{figure*}

\vspace{7pt} \noindent
\textbf{Practical forgetting.}
We consider two more practical and difficult settings, \ie, \textit{few-shot setting} and \textit{missing class setting}.
We conduct single-step and continual forgetting under a few-shot setting where each class has only 4 samples on face recognition tasks.
The results are shown in \cref{tab:single-face,tab:cl-face}.
\cref{fig:cl-obj-few} shows the results for few shot continual forgetting on object detection tasks, where the shot number is 32.
We can find that most baseline methods \textit{degrade heavily} or even fail to forget selective classes while ours can still achieve satisfactory results.
Comparing GS-LoRA and GS-LoRA++, we can find that adding prototype loss can effectively avoid the overfitting of training samples and achieve better results.
\begin{figure}[t]
    \centering
    \includegraphics[width=0.9\columnwidth]{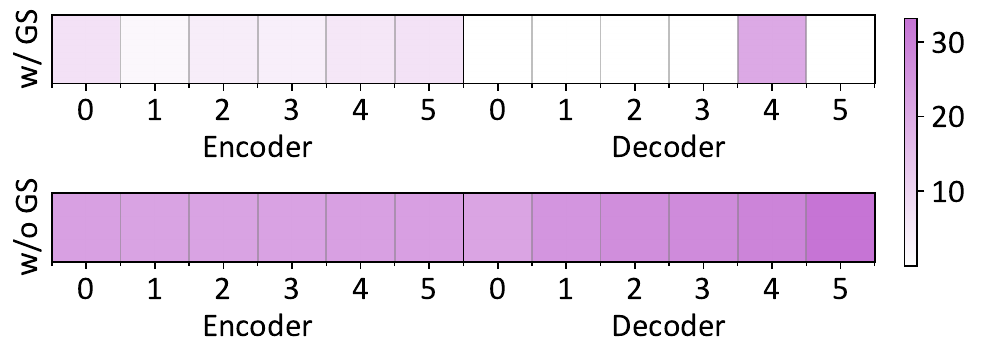}
    \vspace{-3pt}
    \caption{{Comparison of $\ell_2$ norm of each LoRA group with or without group sparse loss.}
    Lighter colors mean smaller $\ell_2$ norms which indicate less model modification.
    The first row shows the result with group sparse loss and the second row is the result of not using it (\ie $\alpha=0$).}
    \label{fig:detr-norm}
    \vspace{-10pt}
\end{figure}
In terms of missing class settings, we perform continual forgetting experiments on face recognition tasks \revise{and image classification tasks}.
Twenty classes are forgotten per task, while five of the remaining classes lack samples, \ie, they do not have replay data.
The results are in \cref{tab:openset}.
\revise{In image classification experiments, we utilize pre-trained ViT weights from PyTorch~\cite{paszke2019pytorch}, initially pre-trained on ImageNet-1k.
In addition to reporting the performance of ImageNet-100, we also report the performance of classes beyond ImageNet-100, \textit{i.e.}, missing class accuracy.
}

We can find that by constraining the number of learnable parameters, our method alleviates forgetting in the classes without replay data dramatically and performs a good performance in the remaining class.
These experiments show the \textit{effectiveness} and \textit{robustness} of GS-LoRA++ when the forgotten samples are rare or some remaining data is missing, which makes it more practical.

\subsection{Ablation Study}
In this part, we conduct comprehensive ablation studies to analyze the effectiveness and efficiency of GS-LoRA++.
If not specified, the default configuration for deformable DETR includes six encoders and six decoders, and the Face Transformer utilizes six Transformer blocks.
The rank we use in GS-LoRA++ is 8, and 10 classes are forgotten in face recognition and object detection tasks.

\begin{table}[t]
    \centering
    \caption{Ablation study of sparse intensity under few-shot settings on Face Transformer.}
    \vspace{-5pt}
     \setlength{\tabcolsep}{11pt}
    \begin{tabular}{lrrrc}
    \toprule 
       $\alpha$ & $Acc_f \downarrow$& $Acc_r \uparrow$& $H \uparrow$ &\begin{tabular}[c]{@{}c@{}}Zero Group\\ Ratio\end{tabular} \\ \midrule
     0 & 10.44 & 71.85 & 66.74 & 0.00 \\
0.0001 & 11.83 & 71.78 & 65.90 & 0.00 \\
0.0005 & 11.60 & 71.95 & 66.11 & 0.00 \\
0.001 & 9.74 & 71.71 & 67.07 & 0.00 \\
\rowcolor{Light}
0.01 & 3.94 & 70.16 & 69.47 & 0.50 \\
0.05 & 1.62 & 69.35 & 70.22 & 0.67 \\
0.1 & 0.00 & 69.72 & 71.20 & 0.83 \\
0.2 & 0.00 & 69.08 & 70.86 & 0.83 \\
0.5 & 6.15 & 67.37 & 66.98 & 0.83 \\
    \bottomrule
    \end{tabular}
    \label{tab:alpha}
\end{table}
\begin{figure}[t]
    \centering
    \includegraphics[width=0.9\columnwidth]{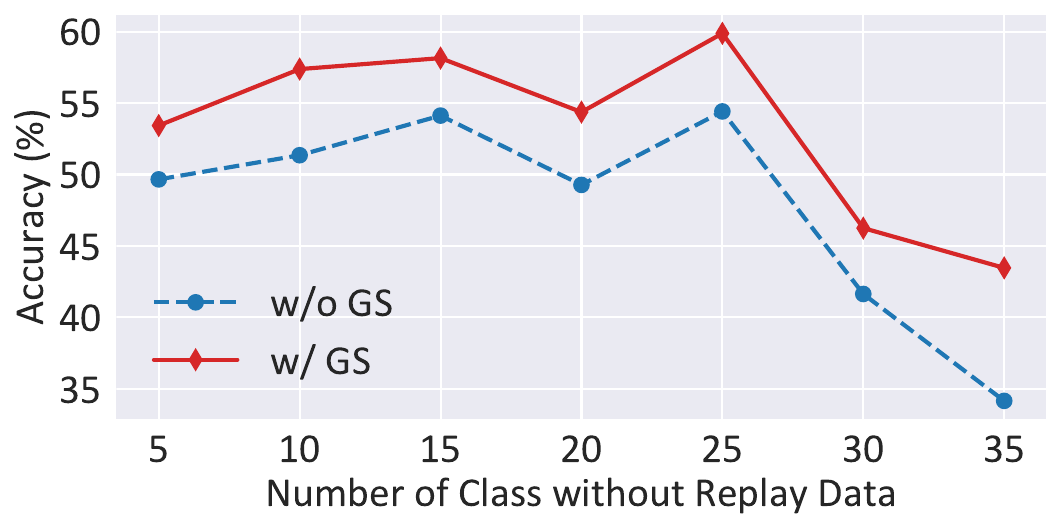}
    \caption{{Ablation study on group sparse (GS) regularization.}
In this experiment, 30 classes are forgotten. Among the remaining 70 classes, only some classes can be replayed. 
The x-axis represents the number of classes \textit{without} replay data, while the y-axis denotes the accuracy of these classes.
}
    \label{fig:open}
\end{figure}

\vspace{7pt}
\noindent
\textbf{Group Sparsity Loss.}
We use group sparse regularization to achieve a sparser and more accurate modification.
In \cref{fig:detr-norm}, we illustrate each parameter group's $\ell_2$  norm in the deformable DETR when forgetting one class.
We can find that our GS-LoRA++ achieves comparable performance with directly using LoRA to fine-tune all FFNs ({forget AP: 0.40 \vs~0, remain AP: 44.49 \vs~44.51}) while requiring to modify significantly \textit{fewer parameters}. 
Meanwhile, we can easily locate the knowledge more precisely with the help of the group sparsity selection strategy. 
The upper layers in the decoder contain more class-specific knowledge and need more modifications.
We also ablate the sparse intensity on Face Transformer under few-shot settings in \cref{tab:alpha}, we can find that with a bigger $\alpha$, the model achieves sparser modification and alleviates the overfitting.
However, using too much sparse intensity ($\alpha=0.5$) can affect the performance. 

If the data of some remaining classes cannot be replayed, \ie, under missing data settings, GS-LoRA++ can effectively reduce catastrophic forgetting in these classes.
We conduct the following experiments on Face Transformer.
Before forgetting, the model can identify 100 people and we want the model to forget 30 people.
\cref{fig:open} shows the results when data of certain remaining classes are not available for replay.
It is clear that our method mitigates catastrophic forgetting in remaining classes.

\vspace{7pt}
\noindent
\textbf{Parameter Efficiency.}
By adjusting the rank of LoRA, we can easily control the learnable parameters.
Here, we study how the rank of LoRA affects the performance when 10 classes are forgotten in the deformable DETR model.
The results in \cref{tab:rank} reveal that a larger rank tends to achieve better performance, but it will also introduce more tunable parameters.
The performance plateaus after rank goes beyond 8.
Remarkably, forgetting can be achieved using only less than 1\% of the parameters with a rank of 8.
However, most continual learning and machine unlearning methods need to modify nearly all parameters, posing inefficiency for large models.

\begin{table}[t!]
\centering 
\caption{{{Ablation study of the rank of LoRA modules.}}
Effective forgetting can be achieved by modifying less than 1\% parameters. }
\vspace{-5pt}
\setlength{\tabcolsep}{8pt}
\renewcommand{\arraystretch}{1.1}
\begin{tabular}{lrrrr}
\toprule
Rank & \begin{tabular}[c]{@{}c@{}}Tunable\\ Ratio\end{tabular}  & $AP_f \downarrow$ & $AP_r \uparrow$ & $H \uparrow$ \\ \midrule
Pre-train & -  & 43.92 & 44.73 & - \\
2 & 0.15\% & 7.60 & 44.45 & 37.98 \\
4 &  0.31\% & 6.76 & 44.33  & 40.42 \\
\rowcolor{Light}
8 & 0.62\% & 3.22 & 44.32 & 42.38 \\
16 & 1.23\% & 3.00 & 44.31 & 42.53 \\
32 & 2.47\%  & 3.26 & 44.32  & 42.40 \\ \bottomrule
\end{tabular}
\label{tab:rank}
%\vspace{-5pt}
\end{table}

\begin{table}[t]
\centering
\caption{{Data efficiency comparison.} Data ratio means the ratio of data used for forgetting to data used for pre-training.}
\vspace{-5pt}
\renewcommand{\arraystretch}{1.1}
\setlength{\tabcolsep}{7pt}
\begin{tabular}{llrrrc}
\toprule
\begin{tabular}[c]{@{}c@{}}Data Ratio\end{tabular} & Speed &$Acc_f \downarrow$ & $Acc_r \uparrow$ & $H \uparrow$ & \begin{tabular}[c]{@{}c@{}}Zero Group\\ Ratio\end{tabular} \\ \midrule
Pre-train & - & 72.74 & 73.81 & - &-\\
0.5 & 2$\times$ & 0.00 & 70.01 & 71.35 & 0.87 \\
0.2 & 5$\times$ & 0.00 & 70.09 & 71.39 & 0.87 \\
\rowcolor{Light}
0.1 & 10$\times$ & 0.00 & 71.35 & 72.04 & 0.50 \\
0.05 & 20$\times$ & 0.00 & 71.71 & 72.22 & 0.50\\
0.02 & 50$\times$ &6.50 &69.47 &67.82 & 0.17\\
\bottomrule
\end{tabular}
\label{tab:ratio}
%\vspace{-5pt}
\end{table}

\begin{table}[t]
    \centering
    \caption{{Ablation study of the shot number of few-shot settings for face recognition task.}
    GS-LoRA++ can work well in most cases, only suffers from performance degradation in particularly extreme cases.}
    \vspace{-5pt}
    \renewcommand{\arraystretch}{1.1}
    \setlength{\tabcolsep}{5.8pt}
    \begin{tabular}{l|rrr>{\columncolor{Light}}rrr}
    \toprule
        Shot Number & Pre-train&1&2&4&8&16\\
        \midrule
        $Acc_f \downarrow$ & 72.74& 52.09&29.70&3.94&0.00&0.00\\
        $Acc_r \uparrow$ &73.81&72.43&71.45&70.16&71.32&70.59 \\
        $H \uparrow$&-&32.14&53.72&69.74&72.02&71.65\\
    \bottomrule
    \end{tabular}
    \label{tab:shot_num}
\end{table}

\vspace{7pt}
\noindent
\textbf{Data Efficiency.}
One benefit of our efficient forgetting paradigm is that we only utilize a small amount of data.
In practical scenarios, using too much data will dramatically increase training costs.
We compare the performance with different data ratios in \cref{tab:ratio}.
Our approach demonstrates satisfactory performance even with minimal training data, which speeds up the forgetting process. 

We further consider the impressive performance of GS-LoRA++ in terms of data efficiency under few-shot settings and conduct ablation studies on the shot number in \cref{tab:shot_num}.
We find our method works well under 4, 8 and 16 shots and only suffers from performance degradation \textit{in particularly extreme cases}.

\begin{figure}[t]
    \centering
    \includegraphics[width=0.8\linewidth]{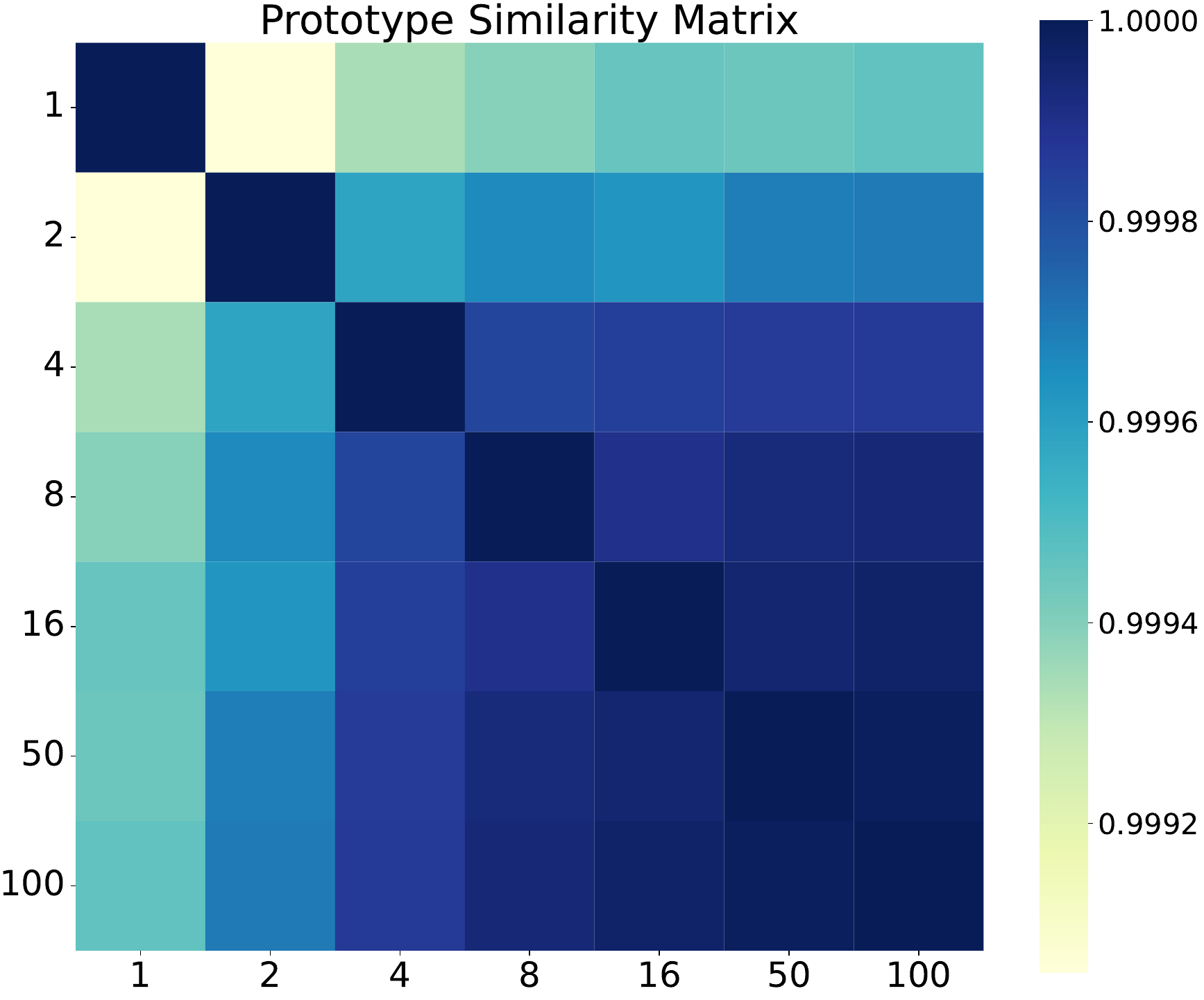}
    \caption{\revise{Prototype similarity between scenarios involving different numbers of shots.}}
    \label{fig:sim}
\end{figure}

\revise{
To explore the reason for failure in extremely limited sample scenarios, we visualize the similarity of prototypes in \cref{fig:sim}.
Here, we define the similarity of prototype sets as follows.
Given two prototype sets $P$ and $P'$, where each set contains the same $c$ classes, the prototype of the $i$-th category in each set is denoted as $p_i$  and $p_i'$, respectively.
The similarity of two prototype sets $S_{P,P'}$ is defined as the average cosine similarity computed over their respective class prototypes, mathematically:
\begin{equation}
    S_{P,P'}=\frac{1}{c}\sum_{i=1}^c cos(p_i,p_i').
\end{equation}
We can find that the similarity between the prototypes from 1/2-shot settings is less similar to the ones from other shot settings (4/8/16/50/100 settings), which leads to critical failures in these cases.
This stems from an inherent statistical constraint: with extremely limited samples, the prototype feature fails to capture robust class distributions, thereby amplifying noise and bias.
Furthermore, to alleviate the bias to some extent, we conduct a simple random augmentation~\cite{randaug} when calculating the prototype and the performance improves in these extreme cases. The experimental results are shown in Tab. \ref{tab:aug}.
}

\begin{table}[h]\color{black}
\centering\small
\setlength{\tabcolsep}{4pt}
 \caption{\revise{Random augmentation~\cite{randaug} on prototype calculation can improve the performance in extreme scenarios~
 (1/2-shot).}}
\begin{tabular}{l|rrrrr}
\toprule
\multirow{2}{*}{\# Shot} & \multirow{2}{*}{pretrain} & \multicolumn{2}{c}{1} & \multicolumn{2}{c}{2} \\  \cmidrule(lr){3-4}  \cmidrule(lr){5-6} 
                             &                           & w/o Aug         & w/ Aug        & w/o Aug         & w/ Aug       \\ \midrule

$Acc_f \downarrow$                            &    72.74                       &  52.09         &   42.46        & 29.70          &   21.69        \\
$Acc_r \uparrow$                            &           73.81                &   72.43        &         72.05  &       71.45    &       66.46    \\
$H \uparrow$                            &                    -       &         32.14  &     42.64      & 53.72          &       57.74    \\ \bottomrule
\end{tabular}
    \label{tab:aug}
\end{table}

\noindent
\textbf{Position of LoRA.}
A lot of studies \cite{meng2022locating,dai2021knowledge,jawahar2019does,geva2020transformer,gueta2023knowledge} have explored the \textit{knowledge} in the language models, but how Transformer models retrieve and utilize this stored knowledge is still an open question \cite{zhang2024comprehensive}.
Besides, few work studies the knowledge in pre-trained \textit{vision models}.
Therefore, we ablate the position of LoRA.
We consider adding LoRA modules on the projection matrices of multi-head attention or FFN modules.
\cref{tab:lorapos} shows the results of different positions of LoRA. 
We can find that editing the multi-head attention or the FFN module can both achieve comparable satisfactory results, indicating the potential to further explore the knowledge of pre-trained vision models.

\begin{table}[t!]
    \centering
    \caption{Ablation study of the position of LoRA.}
    \vspace{-5pt}
    \renewcommand{\arraystretch}{1.1}
    \begin{tabular}{lrrrrc}
    \toprule 
       Pos & \begin{tabular}[c]{@{}c@{}}Tunable\\ Ratio\end{tabular}& $Acc_f \downarrow$& $Acc_r \uparrow$& $H \uparrow$ &\begin{tabular}[c]{@{}c@{}}Zero Group\\ Ratio\end{tabular} \\ \midrule
       Pretrain & - &72.74&73.81&-&-\\
       \rowcolor{Light}
       FFN &1.28\% & {0.00} & {71.35} & 72.04 &0.50 \\
       Multi-Head &0.08\% &0.00 & 69.67 & 71.17 &0.33\\
    \bottomrule
    \end{tabular}
    \label{tab:lorapos}
\end{table}

%% file: sec/6_discussion.tex
\section{Discussion}
\begin{figure}[t]
    \centering
    \includegraphics[width=\linewidth]{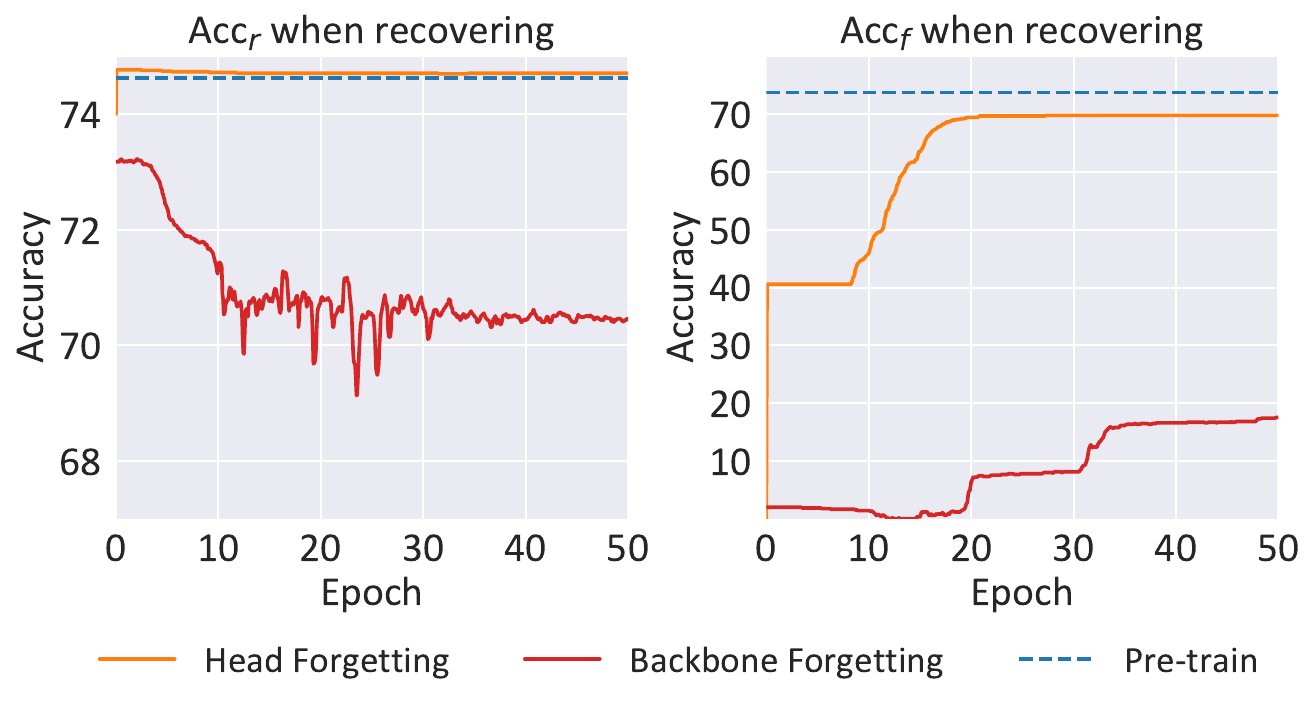}
    \caption{{Accuracy on forgotten classes and retained classes when recovering.} The \textcolor{myblue}{blue} line (Pre-train) is the result before forgetting. The \textcolor{myorange}{orange} line (Head Forgetting) is the trivial masking method. The \textcolor{myred}{red} line (Backbone Forgetting) is the GS-LoRA++.}
    \label{fig:backbone}
    \vspace{0pt}
\end{figure}
\textbf{Real Forgetting or Deceptive Forgetting?}
\label{sec:6.1}
\begin{figure}
    \centering
    \includegraphics[width=0.95\linewidth]{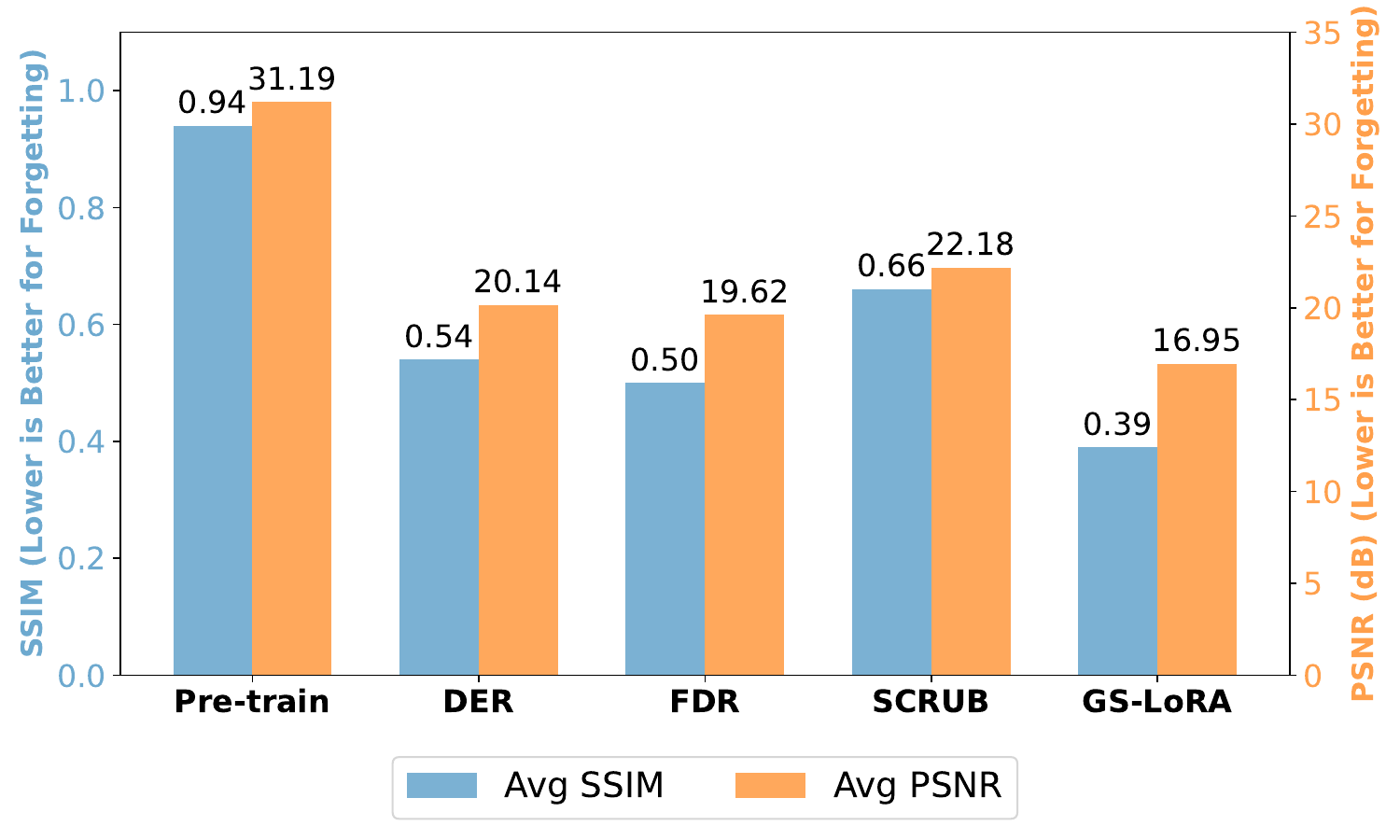}
    \vspace{-7pt}
    \caption{\color{textpurple}Reconstruction attack quality: evidence of real forgetting. Lower ($\downarrow$) values indicate effective removal of privacy/identity information.}
    \label{fig:mia}
    \vspace{-5pt}
\end{figure}
When we want to forget some specific classes, the naive solution is to mask their output FFN weights directly, which we refer to as ``head forgetting'' for simplicity.
However, this trivial solution is \textit{deceptive forgetting} and easy to recover.
It's like a kid who knows the answer and deliberately does not say it.
\textit{Real forgetting} should occur at the backbone and is difficult to recover.

We design the following experiment to demonstrate the significance of backbone forgetting.
We load a model in which forgetting has occurred, freeze its backbone, and fine-tune the output FFN layer using data containing all classes.
Then, we evaluate the model's performance on the forgotten and retained classes.
\cref{fig:backbone} shows the classification accuracy curve with epoch when recovering.
Compared to head forgetting, we can find that the model after forgetting via GS-LoRA++ can only be recovered to approximately 17\% on forgotten classes, significantly lower than  70\% achieved in head forgetting. 
Although it is possible to recover the accuracy of forgotten classes to a very low level in backbone forgetting, such recovery adversely impacts the accuracy of the remaining classes. 
Additionally, the recovery process for our method needs more epochs, while head forgetting can be recovered within 20 training epochs.

\begin{figure}[t]
    \centering
    \includegraphics[width=0.65\linewidth]{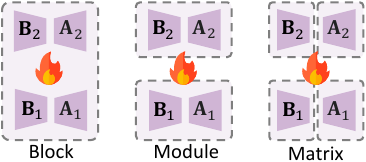}
    \vspace{-5pt}
    \caption{{Illustration of different grouping strategies.}
    The dashed boxes delineate different grouping methods.
}
    \label{fig:grouping}
    %\vspace{-5pt}
\end{figure}

\begin{table}[t]
\centering
\caption{{The scalability of GS-LoRA++} for three Face Transformer with different sizes. 
\# means the number of.}
\vspace{-5pt}
\setlength{\tabcolsep}{1.5pt}
\renewcommand{\arraystretch}{1.1}
\begin{tabular}{cccccccc}
\toprule
\multirow{2}{*}{\# Blocks} & \multirow{2}{*}{\# Param} & \multicolumn{2}{c}{$Acc_f \downarrow $} & \multicolumn{2}{c}{$Acc_r \uparrow $} & \multirow{2}{*}{$H \uparrow$} & \multirow{2}{*}{\begin{tabular}[c]{@{}c@{}}\ Zero Group\\ Ratio \end{tabular}} \\  \cmidrule(lr){3-4} \cmidrule(lr){5-6}
  &  & Pre-train & Forget & Pre-train & Forget &  &  \\ \midrule
 6& 19M &71.46 & 0.00 & 73.18 & 70.62 & 71.04& 0.50 \\
  12& 38M & 76.91 & 0.00 & 76.46 & 71.56 & 74.14&0.92  \\
 18& 57M & 72.74 & 0.00 & 73.81 & 71.35 & 72.04 &  0.78\\ \bottomrule
\end{tabular}
\label{tab:scale}
% \vspace{-1pt}
\end{table}

\begin{table}[t!]
\centering 
\caption{{Effect of grouping strategies.}
The zero-group ratio goes up when using a more detailed grouping strategy.}
\vspace{-5pt}
\setlength{\tabcolsep}{7pt}
\renewcommand{\arraystretch}{1.1}
\begin{tabular}{lcccc}
\toprule
\begin{tabular}[c]{@{}c@{}}Grouping\\ Strategy\end{tabular} & $Acc_f \downarrow$ & $Acc_r \uparrow$ & $H \uparrow$ & \begin{tabular}[c]{@{}c@{}}Zero Group\\ Ratio\end{tabular}  \\ \midrule
Block & 0.00 & 71.35 & 72.04 & 0.50 \\
Module & 0.00 & 70.98 & 71.85 & 0.75 \\
Matrix & 0.00 & 70.09 & 71.39 & 0.87 \\ \bottomrule
\end{tabular}

\label{tab:group}
\vspace{-5pt}
\end{table}

\revised{To further validate that our method achieves ``real forgetting'' rather than ``deceptive forgetting'', we go beyond classification accuracy and recovery experiments to directly probe the information retained within the model's internal representations. A model might learn to suppress a class output while still retaining identifiable features, which could be exploited by privacy attacks. To test this, we conduct a feature reconstruction attack, attempting to reconstruct original images from the forgotten classes using only the feature embeddings from the unlearned model. We employ a method inspired by Deep Image Prior~\cite{ulyanov2018deep} to perform this reconstruction.}

\revised{We quantitatively measure the reconstruction quality using the Structural Similarity Index (SSIM) and Peak Signal-to-Noise Ratio (PSNR)~\cite{hore2010image}, where lower scores indicate poorer reconstruction and thus more effective erasure of information. As illustrated in \Cref{fig:mia}, we compare GS-LoRA against the pre-trained model and several strong baselines (DER, FDR, SCRUB). While the pre-trained model allows for high-quality reconstructions (SSIM: 0.94, PSNR: 31.19 dB), demonstrating the presence of knowledge, GS-LoRA achieves the lowest reconstruction quality scores among all methods (SSIM: 0.39, PSNR: 16.95 dB). This significant degradation provides strong evidence that GS-LoRA does not merely mask outputs but fundamentally alters the model's internal features, making it substantially more difficult to recover private or class-specific information. This supports our claim of achieving a more genuine and robust form of forgetting.}

\vspace{7pt}
\noindent
\textbf{Scalability.} We demonstrate the scalability of GS-LoRA++ in pre-trained models of different sizes.
We first pre-train three Face Transformer models comprising 6 blocks, 12 blocks, and 18 blocks. 
It should be noted that the size of our dataset is limited and slight overfitting occurs when there are 18 blocks.
Then we use GS-LoRA++ to forget selective classes.
As depicted in \cref{tab:scale}, GS-LoRA++ exhibits remarkable scalability, demonstrating effective performance across both large and small models.
Combined with small tunable parameters and high data efficiency, GS-LoRA++ can be a useful tool for privacy erasure in large models in practice.

\vspace{7pt}
\noindent
\textbf{Different Grouping Strategies.}
By default, we regard two LoRA modules in a Transformer block as a group, as illustrated in \cref{fig:pipeline}.
Here, we explore the effectiveness of using GS-LoRA++ with different grouping strategies.
In the FFN module \cite{vaswani2017attention}, there are two linear layers, each of which can add a LoRA module.
And in a LoRA module \cite{hu2021lora}, there are two low-rank matrices.
We consider three grouping strategies: \textit{Block}, \textit{Module}, and \textit{Matrix}, as depicted in \cref{fig:grouping}.
\textit{Block} is the default setting.
\textit{Module} denotes each LoRA \textbf{module} is a group, resulting in twice the number of groups compared to the Transformer blocks.
\textit{Matrix} means each \textbf{matrix} in LoRA modules is a group, and the number of groups is four times the number of Transformer blocks.
The results are in \Cref{tab:group}. With more detailed grouping strategies, the model can achieve excellent performance with sparser modifications.

%% file: sec/7_conclusion.tex
\section{Conclusion}

This paper presents a new and practical problem called continual forgetting and proposes an efficient and effective method to solve it.
For each continual forgetting task, we add a series of LoRA modules and only fine-tune them to achieve knowledge erasure.
Additionally, we adopt a group sparse selection strategy to select specific LoRA groups, which can make the modification more accurate and sparser.
To face the challenges in practical scenarios, we utilize the prototype information as supervision, which can effectively alleviate overfitting under few-shot scenarios.
Thorough experiments demonstrate that our method can achieve effective forgetting under various practical scenarios.
We hope our work will inspire future research directions for continual learning and machine unlearning within the community.

\revise{
\noindent\textbf{Future work.} 
Looking ahead, several promising research avenues can build upon the foundation of GS-LoRA and GS-LoRA++. A critical next step is to conduct a rigorous evaluation of our framework against sophisticated membership inference attacks (MIAs)~\cite{hu2022membership,shokri2017membership,carlini2022membership}. Such an assessment would empirically validate and formally quantify the data privacy protection capabilities of our approach.
From an application perspective, while the current work successfully demonstrates class-wise continual unlearning, a significant extension is the development of instance-wise unlearning~\cite{cha2023learning}. This presents a compelling challenge, as it requires novel strategies for defining "forgetting" and "retaining" sets at the individual data point level.
Fundamentally, we believe the core principle of our approach, using sparse, low-rank adaptations to precisely modify model behavior, is uniquely suited for this granular task. This mechanism could be leveraged not just to forget, but to finely misclassify or alter the learned representation of specific data points, rather than entire classes.
In summary, these future directions, from robust privacy validation to achieving instance-level granularity, are essential for advancing the practical utility and trustworthiness of machine unlearning. Successfully tackling these challenges will not only solidify the contributions of GS-LoRA++ but also pave the way for more agile, controllable, and responsible AI systems.
}